\documentclass{article}
\usepackage[utf8]{inputenc}
\usepackage{geometry}
\usepackage{titlesec}
\usepackage{titletoc}
\usepackage{graphicx}
\usepackage{subfigure} 
\usepackage{float} 
\usepackage{amsmath} 
\usepackage{amssymb} 
\usepackage{mathrsfs} 
\usepackage{listings} 
\usepackage{graphicx} 
 \usepackage{epstopdf}
 \usepackage{bbm}
 \usepackage{bm}
 \usepackage{comment}
 \usepackage{makecell}
\usepackage{algpseudocode}
\usepackage{xcolor}
\usepackage{tikz}
\usepackage[linesnumbered,ruled,vlined]{algorithm2e}
\usepackage{pifont}
\usepackage{appendix}
\DeclareUnicodeCharacter{0301}{*************************************}

\newcommand*\circled[1]{\tikz[baseline=(char.base)]{
            \node[shape=circle,draw,inner sep=2pt] (char) {#1};}}
\SetCommentSty{mycommfont}

\SetKwInput{KwInput}{Input}                
\SetKwInput{KwOutput}{Output} 
\SetKwInput{KwInitialize}{Initialize} 

\def\E{\mathbb{E}}

\def\R{\mathbb{R}}
\def\p{\mathbf{p}}

\newtheorem{theorem}{Theorem}
\newtheorem{lemma}{Lemma}

\newtheorem{problem}{Problem}

\usepackage[
backend=biber,
style=alphabetic,
]{biblatex}
\title{Improved Algorithms for Adversarial Bandits with Unbounded Losses}
\date{}
\author{%
    Mingyu Chen \\
    \scalebox{1}{Boston University}\\
    \scalebox{1}{\texttt{mingyuc@bu.edu}} 
    \and 
    Xuezhou Zhang \\
    \scalebox{1}{Boston University} \\ 
    \scalebox{1}{\texttt{xuezhouz@bu.edu}} 
}

\begin{document}
\maketitle

\begin{abstract}
We consider the Adversarial Multi-Armed Bandits (MAB) problem with unbounded losses, where the algorithms have no prior knowledge on the sizes of the losses. We present \texttt{UMAB-NN} and \texttt{UMAB-G}, two algorithms for non-negative and general unbounded loss respectively. For non-negative unbounded loss, UMAB-NN achieves the first adaptive and scale free regret bound without uniform exploration. Built up on that, we further develop UMAB-G that can learn from arbitrary unbounded loss. Our analysis reveals the asymmetry between positive and negative losses in the MAB problem and provide additional insights. We also accompany our theoretical findings with extensive empirical evaluations, showing that our algorithms consistently out-performs all existing algorithms that handles unbounded losses.
\end{abstract}

\section{Introduction}
Multi-armed bandit (MAB) presents a popular online learning framework for studying decision making under uncertainty \cite{slivkins2019introduction, lattimore2020bandit, bubeck2012regret}, with a wide range of applications such as advertisement \cite{schwartz2017customer}, medical treatments \cite{villar2015multi}, and recommendation systems \cite{mary2015bandits}.
In this paper we focus on the adversarial MAB (AMAB), where the losses are allowed to be generated adversarially by the environment \cite{auer2002nonstochastic}.
Most prior works on AMAB assume that the losses are naturally bounded, e.g. $\ell_t\in[0,1],\forall t$.
With such knowledge, the algorithms can set their \textit{learning rate} (in a general sense) properly.
For example, in its regret analysis, the \texttt{EXP3} algorithm relies on the inequality $\exp(x)\le 1+x+(e-2)x^2$ to transform exponential terms into quadratic terms \cite{auer2002nonstochastic}, which only holds true if the loss $x$ can be upper bounded by $1$.
In many real-world applications, however, such natural loss bound does not always exist. 
For example, in quantitative trading, the fluctuation of stock prices can differ wildly across time. 
In online market places, the price can vary dramatically for different products. 
If one must give a uniform bound $M$ for the losses across all actions and time, such a bound will likely be loose.
In such cases, existing algorithms will have a regret that scales with $M$, which is suboptimal compared to a guarantee that depends on the actual size of the losses.

Motivated by the above limitation of existing algorithms, we wish to design AMAB algorithms that require no prior knowledge on the scale of the losses and \textit{adaptively} achieves smaller regret when the losses are small in scale. 
In addition, instead of a regret bound that depends on the number of rounds and a (hidden) uniform bound of the losses, we wish to design \textit{data-dependent} algorithms whose regret scales with the actual loss sequence, which is beneficial when the sequence of loss is sparse or when its scale varies across time \cite{wei2018more, bubeck2018sparsity}.
In other words, we would like to ask the following question:
\begin{center}
    {Can we design an algorithm that achieves \textbf{optimal} and \textbf{adaptive} regret guarantee \\ \textbf{without} any prior knowledge on the losses?}
\end{center}

\begin{table}[t]
\centering
\resizebox{1\textwidth}{!}{

\begin{tabular}{||c|c|c|c||} 
 \hline
 \textbf{Algorithm} & \textbf{Unbounded}  & \textbf{Adaptive} & \textbf{Regret}  \\ [0.5ex] 
 \hline\hline
 \cite{hazan2011better} & No  & Yes & $\widetilde O\Big(\sqrt{\sum_{t=1}^T\|\ell_t\|^2_{2}}\Big) $\\ 
 \cite{hadiji2020adaptation} & Yes  & No & $\widetilde O\Big(\ell_{\infty}\sqrt{nT}\Big)$\\ 
 \cite{putta2022scale} Non-Adaptive & Yes  & No & $\widetilde O\Big(\ell_{\infty}\sqrt{nT}+\sqrt{n \sum_{t=1}^T \|\ell_t\|^2_2}\Big)$\\
 \cite{putta2022scale} Adaptive & Yes  & Yes & $\widetilde O\Big(\ell_{\infty}\sqrt{n\sum_{t=1}^T \|\ell_t\|_1}+\sqrt{n \sum_{t=1}^T \|\ell_t\|^2_2}\Big)$\\
 \textbf{\texttt{UMAB-G} Non-Adaptive} & Yes  & No & $\widetilde O\Big( \ell_\infty^- \sqrt{nT}  +\sqrt{n \sum_{t=1}^T \|\ell_t\|_\infty^2}\Big)$\\
 \textbf{\texttt{UMAB-G} Adaptive} & Yes  & Yes & $\widetilde O\Big(\ell_{\infty}\sqrt{n \sum_{t=1}^T \|\ell_t\|_\infty}+\sqrt{n \sum_{t=1}^T \|\ell_t\|_\infty^2}\Big)$\\
 [1ex] 
 \hline
\end{tabular}
}
\caption{Comparison between our results and previous works\protect\footnotemark}
\label{table::1}
\end{table}
\footnotetext{For brevity we consider $n, \ell_\infty \ll T$ and omit the log terms. Detailed regret is provided later.}

In the following, we present two algorithms, \texttt{UMAB-NN} and \texttt{UMAB-G}, for \underline{N}on-\underline{N}egative and \underline{G}eneral unbounded loss, respectively.
Our main contributions can be summarized as follows.
\begin{enumerate}
    \item We propose \texttt{UMAB-NN}, a \textbf{scale-free} AMAB algorithm that works for unbounded non-negative losses. The regret guarantee of \texttt{UMAB-NN} adapts to the infinity norm of the loss sequence while matching the worst-case lower bound of \cite{auer2002nonstochastic}.
    \item Building upon \texttt{UMAB-NN}, we then propose \texttt{UMAB-G} which works for arbitrary unbounded losses that can be both possible and negative.
    We present two versions of the algorithm, distinguished by whether the exploration subroutine adapts to the observed losses.
    For the non-adaptive version, it achieves an optimal worst-case regret guarantee and partially adapts to the non-negative part of the loss sequence, improving upon the previous results of \cite{hadiji2020adaptation,putta2022scale, huang2023banker}.
    For the adaptive version, our algorithm achieves an improvement on the order of $\mathcal{O}(\sqrt{n})$ compared to \cite{putta2022scale}, where $n$ is the number of the actions.
    \item 
    Last but not least, we evaluate the performance of our algorithms on real world datasets. 
    The results show that our algorithms consistently outperform existing methods in a variety of tasks with distinct loss patterns.
    We also construct synthetic simulations to illustrate the impact of our exploration strategy and draw comparisons between the two versions of our algorithm. 
\end{enumerate}

\section{Problem Setup and Related Works}
We start with some notations. Let $[n]$ denote the set $\{1,\dots,n\}$ and $[T]$ denote the set $
\{1,\dots, T\}$.
Let $\Delta_n$ be the probability simplex $\{\p\in \R^n: \sum_{k\in [n]}p_k=1; p_k\ge 0, \forall k\in [n]\}$. 
Let $\mathbf{1}_n$ and $\mathbf{0}_n$ be all ones and all zeros $n$-dimensional vector respectively.
Let $\mathbf{e}_k$ denotes the one-hot vector with $1$ on the $k$th entry.
For vectors $\p_t$ and $\ell_{t}$, we use $p_{t,k}$ and $\ell_{t,k}$ to represent the $k$th entry of $\p_t$ and $\ell_t$ respectively.
The L1, L2 and L-infinity norms of $\ell_t$ are denoted as
$
\|\ell_t \|_1 = \sum_{k\in [n]} |\ell_{t,k}|,\   \|\ell_t \|_2 = \sqrt{\sum_{k\in [n]} \ell^2_{t,k}},\   \|\ell_t \|_\infty = \max_{k\in [n]} |\ell_{t,k}|
$
respectively.
We denote by $\ell_\infty = \max_{t\in [T]} \|\ell_t \|_\infty $ the uniform norm bound of the losses.
Moreover, we denote by
$
\ell_\infty^- =  \max_{t\in [T], k\in [n]} |\min(\ell_{t,k},0)|
$
the magnitude of the most negative entry of the losses.
Notice that $\ell_\infty^-\le \ell_\infty$, and $\ell_\infty^-=0$ if the loss sequence is non-negative.
Both $\ell_\infty$ and $\ell_\infty^-$ are unknown to the player through the game.

\paragraph{Adversarial Multi-armed Bandit:} We consider the \textit{oblivious adversarial}
setting.
In each round $t=1,\dots, T$, the player selects a distribution $\p_t$ over $[n]$ and the adversary selects a loss vector $\ell_{t}\in \R^n$ \textit{simultaneously}.
Then, the player samples action $k_t\sim \p_t$ and observes loss $\ell_{t, k_t}$. 
We measure the performance of an algorithm in terms of its \textit{pseudo-regret}:
\begin{align}
\label{Definition::Expected_Regret}
 \mathcal{R}_T:=\E\Big[   \sum\nolimits_{t=1}^T \ell_{t,k_t} -  \min_{k\in [n]}  \sum\nolimits_{t=1}^T \ell_{t,k}\Big]
\end{align}

\subsection{Related Works}


\paragraph{Scale-free algorithms:} Scale-free algorithms are ones whose regret bound scales linearly with respect to $\ell_\infty$, while requiring no knowledge of $\ell_\infty$ a prior \footnote{We note that an alternative and more strict interpretation of scale-free algorithms refers to ones that will not change the sequence of $p_t$'s when losses are multiplied by a positive constant.}.
Scale-free regret bounds were first studied in the full information setting, such as experts problems \cite{freund1997decision, de2014follow, cesa2007improved} and online convex optimization \cite{mayo2022scale, jacobsen2023unconstrained, cutkosky2019artificial}.
For experts problems, the \texttt{AdaHedge} algorithm from \cite{de2014follow} achieves the first scale-free regret bound.
For online convex optimization, past algorithms can be categorized into two generic algorithmic frameworks: Mirror Descent (MD) and Follow The Regularizer Leader (FTRL).
The scale-free regret from the MD family is achieved by \texttt{AdaGrad} proposed by \cite{duchi2011adaptive}.
However, the regret bound of \cite{duchi2011adaptive} is only non-trivial when the Bregman divergence associated with the regularizer can be well bounded.
Later, the \cite{orabona2018scale} proposed the \texttt{AdaFTRL} algorithm which achieves the first scale-free regret bound in the FTRL family and generalizes \cite{duchi2011adaptive}'s results to cases where the Bregman divergence associated with the regularizer is unbounded.
For the AMAB problem, \cite{hadiji2020adaptation} extends the method of \cite{duchi2011adaptive} and provides a scale-free regret bound of $\widetilde O\Big(\ell_{\infty}\sqrt{nT}\Big)$, which is optimal (up to log terms) in the worst case.
However, such worst-case regret bounds can be overly pessimistic in general cases: a single outlier loss $\ell_{outlier}$ can result in an additional regret on the order of $O(\|\ell_{outlier}\|_\infty\sqrt{nT})$.
To address it, \cite{putta2022scale} presents scale-free bounds that adapt to the individual size of losses across time.
Unfortunately, the worst-case guarantee of \cite{putta2022scale} is $\widetilde O\Big(\ell_{\infty}n\sqrt{T}\Big)$, which scales linearly to the number of actions.
Our paper closes this gap: our algorithms achieve an adaptive regret better than \cite{putta2022scale}, as well as an optimal worst-case regret that matches with \cite{hadiji2020adaptation}.

\paragraph{Adaptive algorithms:} Adaptive algorithms refer to the algorithms that dynamically adjusts to the input data it encounters. 
Rather than scaling solely on $T$ in the regret, an adaptive algorithm adapts to a ``measure of hardness'' of the sequence of losses.
An adaptive regret algorithm performs better than the worst-case regret if the sequence of loss is ``good''.
In the last two decades, adaptive algorithms have been widely studied in the settings of expert problems and online convex optimization \cite{hazan2007adaptive,streeter2010less,duchi2011adaptive,de2014follow,orabona2015scale,orabona2018scale}.
For the MAB setting, several works derive adaptive regret bounds based on different ``measure of hardness''.
For example, \cite{allenberg2006hannan, foster2016learning, pogodin2020first, ito2021parameter} derive the first-order regret (a.k.a. \textit{small-loss regret}), which depends on the cumulative loss $\min_{k\in [n]}\sum_{t\in [T]}|\ell_{t,k}|$, but under the assumption that $\ell_{t,k}\in[0,1], \forall t,k$.
\cite{hazan2011better, bubeck2018sparsity, wei2018more, ito2021parameter} propose bounds that depend on the empirical variance of the losses, i.e., $\sum_{t\in [T]}\|\ell_{t}\|_2^2$.
Path-length bounds are also studied \cite{wei2018more, bubeck2019improved, zimmert2021tsallis, ito2021parameter}, which depends on the fluctuation of loss sequence $\sum_{t\in [T]}\|\ell_{t}-\ell_{t-1}\|_1$.
We remark that \textit{all} results above require the assumption that losses are bounded within $[0,1]$, which we remove in this paper.

\section{Algorithm and Analysis}
We now present our two algorithms \texttt{UMAB-NN} and \texttt{UMAB-G}.
\texttt{UMAB-NN} works for the case where losses are \underline{N}on-\underline{N}egative, i.e., $\ell_{t}\in \R_+^n$.
Remarkably, \texttt{UMAB-NN} is a \textit{strictly scale-free} algorithm: the algorithm will not change its sequence of action distributions if the sequence of losses is multiplied by a positive constant, which immediately implies scale-free regret.
Our second algorithm, \texttt{UMAB-G}, builds upon the first algorithm to allow potentially negative losses, i.e., $\ell_{t}\in \R^n$.
We provide two versions of the algorithm: \texttt{UMAB-G} with non-adaptive and adaptive exploration rates.
For the non-adaptive version, our results achieve adaptability to the non-negative part of the loss, while ensuring the optimality for the worst case guarantee, which is new compared to previous works\footnote{We note that a recent work \cite{huang2023banker} proposes an algorithm that claims to achieve adaptive regret for general unbounded loss. However, there exists a critical issue within their proof and algorithm, resulting in their regret being actually unbounded. We have communicated and confirmed with the authors about the issue. More details are provided in Appendix~\ref{Banker_error}.}.
For the adaptive version, we improve the previous result \cite{putta2022scale} by $\mathcal{O}(\sqrt{n})$.
A summary of the comparisons to prior works can be found in Table~\ref{table::1}.

Both the algorithms we propose are based on the Follow-the-Regularized-Leader (FTRL) framework.
Let us first consider the full information case, the traditional adaptive FTRL framework uses a regularizer $\Psi$ and time-varying learning rates $\eta_1,\dots, \eta_{T+1}$, with certain regularity constraints (see, e.g., \cite{orabona2015scale}). 
The update rule takes the form of
\begin{align}
\label{FTRL::update_rule}
    \p_1 = \arg\min_{\p\in \Delta_n} \frac{1}{\eta_1}\Psi(\p),\qquad \p_t = \arg\min_{\p\in \Delta_n}  \Big( \sum_{s=1}^{t-1}\langle  \ell_s, \p \rangle + \frac{1}{\eta_{t }}\Psi(\p) \Big),
\end{align}
where $\ell_s$ is the observed loss at round $s$ and $\eta_{t}$ is the adaptive learning rate depending on the losses $\ell_1,\dots,\ell_{t-1}$.
In the bandit setting, we cannot observe the complete loss vector $\ell_t$.
Similar to prior works, we construct an unbiased loss estimator through the importance-weighted (IW) sampling method introduced by \cite{auer2002nonstochastic}, i.e., construct $\hat\ell_t\in \R^n$ such that
\begin{align*}
    \hat\ell_{t,k} = \frac{\mathbbm{1}(k=k_t)}{p_{t,k}}\ell_{t,k},\ \forall k\in [n],
\end{align*}
where $\mathbbm{1}(k=k_t)$ denotes the indicator function.
Notice that 
\begin{align*}
    \E[\hat\ell_{t}] =  \sum_{k=1}^n p_{t,k} \frac{\mathbf{e}_k}{p_{t,k}}\ell_{t} = \ell_{t}.
\end{align*}
Using $\hat\ell_t$, we are able to reduce the bandit setting to the full information case.

\subsection{Non-negative loss}
Let's start with the setting where the loss sequence is non-negative but can be arbitrarily large, i.e., $\ell_{t,k}\ge 0$ for every $t\in [T]$ and $k\in [n]$.
\texttt{UMAB-NN} (Algorithm~\ref{FTRl::non-negative}) is a natural adaptation of the classic FTRL algorithm with log-barrier regularizer.
The log-barrier regularizer is defined as 
\begin{align*}
    \Psi(\p_t) = \sum_{k=1}^n   \Big(  \log\Big(\frac{1}{p_{t,k}}\Big)  -\log\Big(\frac{1}{n}\Big)    \Big).
\end{align*}
Notice that $\Psi(\p)\ge 0$ for all $\p\in \Delta_n$.
Such regularizers are commonly used for studying adaptive regret in the AMAB setting \cite{wei2018more,putta2022scale,bubeck2019improved}.
In each round, \texttt{UMAB-NN} calculates an action distribution $\p_t$ through the update rule, then plays action $k_t$ sampled from $\p_t$.
After receiving loss $\ell_{t,k}$, \texttt{UMAB-NN} constructs the unbiased IW estimator $\hat\ell_t$ and updates the learning rate $\eta_t$.
The novelty comes in our design of learning rate (line 5). Different from the learning rate in \cite{orabona2018scale}, we use $\ell^2_{t,k_t}$ instead of $\|\hat\ell_t\|_2^2$.
This is because $\|\hat\ell_t\|_2^2$ is of order $1/p^2_{t,k_t}$.
If one uses the one in \cite{orabona2018scale} instead, i.e. $\eta_{t+1} = O( \sqrt{n/\sum_{s=1}^{t} \|\hat \ell_{s}\|_2^2})$, the learning rate will be too small since $1/p^2_{t,k_t}$ cannot be bounded. 
Based on this observation, \texttt{UMAB-NN} adapts the learning rate to the sum of the square of the partial loss, i.e., $\eta_{t+1} = O( \sqrt{n/\sum_{s=1}^{t}  \ell_{s,k_s}^2})$, which can be well bounded by $O(\ell_{\infty}\sqrt{n/T})$.

\begin{algorithm}[t]
\label{FTRl::non-negative}
\DontPrintSemicolon  
  \KwInput{Log-barriers regularization $\Psi$, $\eta_1=\infty$}
 \For{$t=1,\dots,T$}
 {
    Compute the action distribution $\p_t =\arg\min_{\p\in \Delta_n} \Big( \sum_{s=1}^{t-1}\langle \hat \ell_s, \p \rangle + \frac{1}{\eta_{t}}\Psi(\p) \Big)$\;
    Sample and play action $k_t\sim \p_t$. Receive loss $\ell_{t, k_t}$\;
    Construct IW estimator $\hat\ell_t$ such that $\hat \ell_{t,k} =   \frac{\mathbbm{1}(k=k_t)}{p_{t,k}}\ell_{t,k},\ \forall k\in [n] $\;
    Update learning rate $\eta_{t+1} = 2\sqrt{\frac{n}{\sum_{s=1}^{t} \ell_{s,k_s}^2}}$\;
 }
\caption{\texttt{UMAB-NN}: Unbounded AMAB for Non-Negative loss}
\end{algorithm} 
We remark that Algorithm~\ref{FTRl::non-negative} is strictly scale-free. If all losses are multiplied by a constant $c$, then in line 2, both terms on the right hand side will be multiplied by $c$, resulting in the same $p_t$ being picked by the algorithm.
Our main result is the following regret bound for Algorithm~\ref{FTRl::non-negative}.
\begin{theorem}
\label{regret::ALG1}
For any $\ell_1,\dots, \ell_T\in \R_+^n$, the expected regret of Algorithm~\ref{FTRl::non-negative} is upper bounded by 
\begin{align*}
    \mathcal{R}_T \le \tilde{\mathcal{O}}\Big(\sqrt{n\sum_{t=1}^T \|\ell_t\|_\infty^2 } \Big)
\end{align*}
\end{theorem}
Notice that Theorem~\ref{regret::ALG1} is adaptive to the infinity norm of the losses.
Furthermore, the worst case regret is bounded by $\tilde{\mathcal{O}}(\ell_{\infty}\sqrt{nT})$, which matches the lower bound established in \cite{auer2002nonstochastic}. We remark that Theorem \ref{regret::ALG1} is the first result that achieves both optimal adaptive rate and optimal minimax rate for unbounded non-negative losses. Next, we briefly highlight the key steps in proving Theorem \ref{regret::ALG1}, which also provide intuition for our further improvement in the next section.

\paragraph{Proof sketch of Theorem \ref{regret::ALG1}}
Since $\hat\ell_t$ is an unbiased estimator of $\ell_t$ for every $t\in [T]$ and comparator $\p^\dagger\in \Delta_n$, we have
\begin{align*}
  \E\Big[   \sum_{t=1}^T \ell_{t,k_t} -    \sum_{t=1}^T \langle \ell_t,\p^\dagger \rangle \Big] = \E\Big[\sum_{t=1}^T \langle \hat \ell_t, \p_t-\p^\dagger\rangle\Big].
\end{align*}
It suffices to focus on bounding $\sum_{t=1}^T \langle \hat \ell_t, \p_t-\p^\dagger\rangle$.
We start with the standard analysis of a FTRL-type algorithm.
\begin{lemma} (\cite{orabona2019modern} Lemma 7.1)
\label{Lemma::FTRL_regret}
For any $\hat \ell_1,\dots,\hat \ell_T\in \R^n $, using the update rule of (\ref{FTRL::update_rule}) along with the non-increasing sequence of learning rates $\eta_1,\dots,\eta_{T+1}$, there is
    \begin{align*}
    \sum_{t=1}^T \langle \hat \ell_t, \p_t-\p^\dagger \rangle\le \frac{\Psi(\p^\dagger) }{\eta_{T+1}}+ \sum_{t=1}^T \Big( \langle \hat \ell_t, \p_t-\p_{t+1} \rangle+ F_t(\p_t)-F_{t}(\p_{t+1}) \Big)
\end{align*}
for every comparator $\p^\dagger \in \Delta_n$, where function $F_t$ is defined as
\begin{align*}
   F_t(\p) =  \sum_{s=1}^{t-1}\langle \hat \ell_s, \p \rangle + \frac{1}{\eta_{t}}\Psi(\p).
\end{align*}
\end{lemma}
For the sake of completeness, the proof of Lemma~\ref{Lemma::FTRL_regret} is provided in the appendix.
Lemma~\ref{Lemma::FTRL_regret} decomposes the regret into two terms. 
The first term depends on the regularizer and the comparator.
Intuitively, $\Psi(\p^\dagger)$ will appear to be infinity if $\p^\dagger$ is the best fixed action (some entries of $\p^\dagger$ are zeros).
The problem can be easily solved by comparing with some close neighbor of the best
action \cite{putta2022scale}, i.e., mixing a uniform distribution with the best fixed action. 
Therefore, it suffices to focus on the terms $ \langle \hat \ell_t, \p_t-\p_{t+1} \rangle+ F_t(\p_t)-F_{t}(\p_{t+1})$.
The following key lemma gives an upper bound using the notions of local norms.
\begin{lemma}
\label{Lemma::local_norm}
    For any $\hat \ell_1,\dots,\hat \ell_T\in \R^n $, using the update rule of (\ref{FTRL::update_rule}), denote by $\|\mathbf{x}\|_{\mathbf{A}} = \sqrt{\mathbf{x}^\top \mathbf{A}\mathbf{x}}$, there is
    \begin{align}
    \label{bound::general}
        \langle \hat \ell_t, \p_t-\p_{t+1} \rangle+ F_t(\p_t)-F_{t}(\p_{t+1})\le \frac{1}{2}\eta_{t}\| \hat \ell_t\|^2_{(\nabla^2 
 \Psi(\xi_t))^{-1}},
    \end{align}
    where $\xi_t$ is a point between $\p_t$ and $\p_{t+1}$. Moreover, it suffices to set $\xi_t$ as $\p_t$ when $\hat \ell_t\in \R_+^n$.
\end{lemma}
Note that (\ref{bound::general}) holds for general losses and will be useful in the next section. When $\hat \ell_t \in \R_+^n $, 
we can further bound (\ref{bound::general}) by 
$ \min \Big(     \frac{1}{2}\eta_{t} \ell_{t,k_t}^2, |\ell_{t,k_t}| \Big)$, since 
\begin{align}\label{eq4}
    \langle \hat \ell_t, \p_t-\p_{t+1} \rangle+ F_t(\p_t)-F_{t}(\p_{t+1})\le \langle \hat \ell_t, \p_t \rangle = |\ell_{t,k_t}|,
\end{align}
which implies
\begin{align}
\label{bound3}
    \sum_{t=1}^T \langle \hat \ell_t, \p_t-\p^\dagger \rangle\le \frac{\Psi(\p^\dagger) }{\eta_{T+1}}+ \sum_{t=1}^T \min \Big(     \frac{1}{2}\eta_{t} \ell_{t,k_t}^2, |\ell_{t,k_t}| \Big).
\end{align}
The right hand side of (\ref{bound3}) takes a similar form as in scale-free online convex optimization \cite{orabona2018scale}, but the upper bound depends on $\ell_{t,k_t}$ instead of $\|\ell_{t}\|_2$.
Using a learning rate as in Algorithm~\ref{FTRl::non-negative}, the second term on the right hand side of (\ref{bound3}) can be bounded by $\mathcal{O}(\sqrt{n\sum_{t=1}^T\ell_{t,k_t}^2})$ based on \cite{orabona2018scale}, which suffices to complete the proof.


\begin{algorithm}[!t]
\label{FTRl::general}
\DontPrintSemicolon  
  \KwInitialize{Log-barriers regularization $\Psi$, learning rate $\eta_1=1/4$, exploration rate $\rho_1=1/2n^2$, clipping threshold $C_1 =-1$}
 \For{$t=1,\dots,T$}
 {
    Compute the action distribution: $\p_t =\arg\min_{\p\in \Delta_n} \Big( \sum_{s=1}^{t-1}\langle \hat \ell_s', \p \rangle + \frac{1}{\eta_{t}}\Psi(\p) \Big)$\;
    Calculate $\p_t'$ by Algorithm~\ref{FTRl::extra_exploration} with exploration rate $\rho_t$. Play action $k_t\sim \p_t'$. Receive loss $\ell_{t, k_t}$.\;
    Construct loss estimator $\hat\ell_t'$ such that $ \hat \ell_{t,k}' =   \frac{\mathbbm{1}(k=k_t)\ell^\prime_{t,k}}{p_{t,k}'},\ \forall k\in [n]$, where $\ell^\prime_{t,k_t} = \max ( 2C_t,  \ell_{t, k_t}  )$.\;
    Update clipping threshold $C_{t+1} =  \min(C_t, \ell'_{t, k_t})$.\;
    Update learning rate: $ \eta_{t+1} = \frac{1}{4}\sqrt{\frac{n}{nC^2_{t+1}+\sum_{s=1}^{t} {\ell^\prime}_{s,k_s}^2}}$.\;
    Update exploration rate:
    \begin{enumerate}
        \item (Non-Adaptive): $\rho_{t+1} = 1/(2n^2+\sqrt{nT})$.
        \item (Adaptive): $\rho_{t+1} = 1/(2n^2+2\sqrt{\sum_{s=1}^t | \langle \hat\ell_s  , \mathbf{c}_t \rangle|})$.
    \end{enumerate}
 }
\caption{\texttt{UMAB-G}: Unbounded AMAB for General Loss}
\end{algorithm}

\begin{algorithm}[t]
\label{FTRl::extra_exploration}
\DontPrintSemicolon  
 \KwInput{Action distribution $\p_t$. Exploration rate $\rho_t\le 1/2n^2$}
  \KwOutput{Extra exploration distribution $\p_t'$}
    Define $k_t^\star \in  \arg\max_{k'\in [n]} p_{t,k'}$. Construct a vector $\mathbf{c}_t\in \R^n$ such that for every $k\in [n]$, there is
    \begin{align*}
        c_{t,k} = \begin{cases}
			1, & \text{if $p_{t,k}<\rho_t$}\\
            -\sum_{k'\in [n]/\{k\}} c_{t,k'} & \text{if $k = k_t^\star$}\\
            0, & \text{else}
		 \end{cases}
    \end{align*}
    Construct the extra exploration distribution $\p_t' = \p_t+ \rho_t \mathbf{c}_{t}$.
\caption{Extra Exploration on Action Distribution}
\end{algorithm}

\subsection{General loss}
Next, we remove the non-negative assumption and study the general loss setting, i.e., $\ell_1,\dots,\ell_T\in \R^n$.
To begin with, we first explain why Algorithm~\ref{FTRl::non-negative} cannot work when the losses become negative.
Recall Lemma~\ref{Lemma::local_norm}, it requires bounding $\langle \hat \ell_t, \p_t-\p_{t+1} \rangle+ F_t(\p_t)-F_{t}(\p_{t+1})$ by $ \eta_{t}\| \hat \ell_t\|^2_{(\nabla^2 \Psi(\xi_t))^{-1}}/2$ for general losses.
However, notice that 
\begin{align}
\label{bound4}
     \| \hat \ell_t\|^2_{(\nabla^2  \Psi(\xi_t))^{-1}} = \sum_{k=1}^n \frac{\hat \ell_{t,k}^2 }{\nabla^2_{k,k}  \Psi(\xi_t)}  = \sum_{k=1}^n \frac{ \ell_{t,k}^2 \mathbbm{1}(k=k_t) }{p_{t,k}^2} \xi_{t,k}^2 = \frac{ \ell_{t,k_t}^2 }{p_{t,k_t}^2} \xi_{t,k_t}^2,
\end{align}
where $\xi_{t,k_t}$ is some value between $p_{t,k_t}$ and $p_{t+1,k_t}$.
Given $p_{t+1,k_t}$ might significantly exceed $p_{t,k_t}$, the size of $\xi_{t,k_t}/p_{t,k_t}$ cannot be confined.
In this case, ${ \ell_{t,k_t}^2 } \xi_{t,k_t}^2/{p_{t,k_t}^2}$ is potentially of order $O(1/p_{t,k_t}^2)$, which is too large for the analysis.
Additionally, $-\langle\hat\ell_t,\p_{t+1} \rangle$ could potentially be positive and cannot be well bounded due to the same reason, which implies that \eqref{eq4} will not go through.
Thus, inequality (\ref{bound3}) no longer holds under the condition of general loss.
Inspired by such observations, it naturally follows to consider bounding the magnitude of $p_{t+1,k_t}/p_{t,k_t}$.
Unfortunately, without imposing additional restrictions on the losses, using the update (\ref{FTRL::update_rule}) directly cannot bound ${p_{t+1,k_t}}/{p_{t,k_t}}$.
For example, given arbitrary $\p_t$, $\eta_{t+1}$, and $k_t$, we can always find a sufficiently small $\ell_{t,k_t}<0$ that makes $p_{t+1,k_t}\ge 1/2$ through (\ref{FTRL::update_rule}).
In this case, if $p_{t,k_t}$ is close to zero, $p_{t+1,k_t}/p_{t,k_t}$ could be extremely large.

To address this issue, we propose \texttt{UMAB-G}, as illustrated in Algorithm~\ref{FTRl::general}.
The key ideas of \texttt{UMAB-G} include
(1) using truncated loss to update the action distribution.
Instead of directly taking $\hat\ell_t$ as the input loss, we clip it by a threshold $C_t$ that depends on previous received losses $\hat \ell_1,\dots,\hat \ell_{t-1}$.
The truncation ensures that every input loss is ``not too negative'' for the update of action, and thus the magnitude of $p_{t+1,k_t}/p_{t,k_t}$ can be well bounded.
(2) adding an extra exploration to ensure that the probability $p_{t,k}$ would not be overly small.
For unbounded AMAB with general loss, we need to ensure that each arm has a certain probability to be pulled, so that we can perceive the change of loss norm in time to tune the learning rate.
Instead of the commonly used scheme of mixing with a uniform distribution \cite{hadiji2020adaptation, putta2022scale}, we develop a data-dependent mixing strategy (Algorithm~\ref{FTRl::extra_exploration}) that substantially reduces the error caused by the extra exploration.
Specifically, similar to \cite{putta2022scale}, we consider two exploration rate distinguished by whether the exploration rate is adaptive.
The main result of Algorithm~\ref{FTRl::general} is as follows. 
\begin{theorem}
    \label{regret:ALG2}
    For any $\ell_1,\dots,\ell_T\in \R^n$, with the non-adaptive and adaptive exploration rate, the expected regret of Algorithm~\ref{FTRl::general} is upper bounded by
        \begin{align}
             &\text{Non-Adaptive:} &\mathcal{R}_T &\le \tilde{\mathcal{O}}\Big( \ell_\infty n^2+\sqrt{n\sum_{t=1}^T\|\ell_t\|^2_\infty}+\ell_\infty^{-} \sqrt{nT} \Big), \label{regret:ALG2:non_adaptive}\\
             & \text{Adaptive:}   &\mathcal{R}_T &\le \tilde{\mathcal{O}}\Big( \ell_\infty n^2+ \sqrt{n\sum_{t=1}^T \|\ell_t\|_\infty^2 } +\ell_\infty \sqrt{n\sum_{t=1}^T \|\ell_t \|_{\infty}}+\sqrt{n\sum_{t=1}^T\|\ell_t \|_{\infty}} \Big) \label{regret:ALG2:adaptive}
        \end{align}
\end{theorem}
Notice that the non-adaptive regret in Theorem~\ref{regret:ALG2} achieves ``semi-adaptivity'' to the loss sequence.
If the loss sequence is non-negative, the right hand side of (\ref{regret:ALG2:non_adaptive}) is reduced to a form of the regret in Theorem~\ref{regret::ALG1}.
Moreover, the worst case bound of (\ref{regret:ALG2:non_adaptive}) is $\tilde{\mathcal{O}}(\ell_{\infty}\sqrt{nT})$ for large $T$, which is optimal up to log factors \cite{auer2002nonstochastic}.
For the adaptive exploration rate, our result improves upon the previous result \cite{putta2022scale} and achieves optimal dependency on $n$ and $T$.

\paragraph{Proof sketch of Theorem \ref{regret:ALG2}}
Recall that $\hat\ell_t$ is the unbiased estimator and $\hat\ell_t'$ is the clipping biased estimator.
By Algorithm~\ref{FTRl::general} and the proof of Theorem~\ref{regret::ALG1}, it suffices to bound the expectation of $\sum_{t=1}^T \langle \hat\ell_{t}, \p_t'-\p^\dagger \rangle$.
We first decompose the regret into three terms as follows.
\begin{align*}
    \sum_{t=1}^T \langle \hat\ell_{t}, \p_t'-\p^\dagger \rangle = 
    \underbrace{\sum_{t=1}^T \langle \hat\ell_{t}', \p_t-\p^\dagger \rangle}_{ \circled{1} }+
    \underbrace{\sum_{t=1}^T \langle \hat\ell'_{t}, \p_t'-\p_t \rangle}_{ \circled{2}}+
    \underbrace{\sum_{t=1}^T \langle \hat\ell_{t}-\hat\ell_{t}', \p_t'-\p^\dagger \rangle}_{ \circled{3}}.
\end{align*}
Here, term $\circled{1}$ is the regret of the corresponding FTRL algorithm with truncated loss $\hat\ell_1',\dots, \hat\ell_T'$.
Term $\circled{2}$ measures the error incurred by extra exploration, i.e., using $\p_t'$ instead of $\p_t$.
Term $\circled{3}$ corresponds to the error of using the truncated loss $\hat\ell_t'$.
In the rest of the proof, we bound these three terms respectively.

\textbf{Bounding $\circled{1}$}:
By Lemma~\ref{Lemma::FTRL_regret} and Lemma~\ref{Lemma::local_norm}, we have 
\begin{align*}
    \sum_{t=1}^T \langle \hat\ell_{t}', \p_t-\p^\dagger \rangle &\le \frac{\Psi(\p^\dagger) }{\eta_{T+1}}+ \frac{1}{2} \sum_{t=1}^T \eta_{t}\| \hat \ell_t'\|^2_{(\nabla^2 
 \Psi(\xi_t))^{-1}}= \frac{\Psi(\p^\dagger) }{\eta_{T+1}}+ \frac{1}{2} \sum_{t=1}^T \eta_{t} {\ell'}_{t,k_t}^2 \frac{ {p}_{t,k_t}^2   }{   {p'}_{t,k_t}^2  } \frac{\xi_{t,k_t}^2}{{p}_{t,k_t}^2 }.
\end{align*}
The key step is to bound the magnitude of ${p}_{t,k_t}/{p'}_{t,k_t}$ and ${p}_{t+1,k_t}/{p}_{t,k_t}$ (since $\xi_{t,k_t}$ is always between ${p}_{t,k_t}$ and ${p}_{t+1,k_t}$) for $\ell_{t,k_t}\le 0$.
This in turn is guaranteed by our design of loss truncation and extra exploration, which is illustrated in the following lemma.
\begin{lemma}
 \label{Lemma::Proof::3}
     Given any action sequence $k_1,\dots,k_T$, if $\ell_{t,k_t}\le 0$. there is $ {p}_{t,k_t}\le 2 p'_{t,k_t}$ and ${p}_{t+1,k_t}\le 6 {p}_{t,k_t}$ for every $t\in [T]$.
 \end{lemma}
Lemma~\ref{Lemma::Proof::3} ensures that both $ {p}_{t,k_t}/p'_{t,k_t}$ and ${p}_{t+1,k_t}/ {p}_{t,k_t}$ can be bounded by constants. 
With these two ratio bounded, we can immediately reduce the right-hand-side to the form of (\ref{bound3}).
Using a similar proof as in Section 3.1, we can bound $\circled{1}$.


\textbf{Bounding $\circled{2}$}:
By the definition of $\p_t'$, we first note that $ \sum_{t=1}^T \langle \hat\ell'_{t}, \p_t'-\p_t \rangle = \sum_{t=1}^T {\rho_t} \langle \hat\ell'_{t}, \mathbf{c}_t \rangle$, where $\rho_t$ is the exploration rate and $\mathbf{c}_t$ is an offset on $p_t$ to prevent some entries in action distribution from being too small.
The key of our extra exploration algorithm is to upper bound $\langle \hat\ell'_{t}, \mathbf{c}_t \rangle$ by $\mathcal{O}(\ell_\infty \sqrt{nT})$, in contrast to $\mathcal{O}(\ell_\infty n^{3/2}\sqrt{T})$ as in \cite{putta2022scale}.
This reduces the variance of our exploration rate, leading to an improved regret. 
The details are provided in Lemma~\ref{Lemma::Proof::4} as follows.

\begin{lemma}
 \label{Lemma::Proof::4}
 With the non-adaptive and adaptive exploration rates as in Algorithm~\ref{FTRl::extra_exploration}, we have
     \begin{align*}
      &\text{Non-Adaptive:}\    &\E\Big[\sum_{t=1}^T \langle \hat\ell'_{t}, \p_t'-\p_t \rangle\Big] &\le 2\sqrt{n \sum_{t=1}^T \|\ell_t\|^2_{\infty} },\\
      &\text{Adaptive}\  &\E\Big[\sum_{t=1}^T \langle \hat\ell'_{t}, \p_t'-\p_t \rangle\Big] &\le  2n^2\ell_{\infty} + 2\sqrt{1+4n \sum_{t=1}^T \|\ell_t \|_{\infty} }+2\ell_{\infty}\sqrt{n \sum_{t=1}^T \|\ell_t \|_{\infty} }.
     \end{align*}
 \end{lemma}


\textbf{Bounding $\circled{3}$}:
Notice that
\begin{align*} 
    \sum_{t=1}^T \langle \hat\ell_{t}-\hat\ell_{t}', \p_t'-\p^\dagger \rangle\le \sum_{t=1}^T  \|  \hat\ell_{t}-\hat\ell_{t}'\|_1\|\p_t'-\p^\dagger\|_{\infty} 
    \le  \sum_{t=1}^T  \|  \hat\ell_{t}-\hat\ell_{t}'\|_1.
\end{align*}

The key idea of bounding $\circled{3}$ is to show that the number of distinct $(\hat\ell_{t},\hat\ell_{t}')$ pairs and $\|\hat\ell_{t}\|_\infty$ can be bounded by $\mathcal{O}(\log \ell_\infty)$ due to the double tricks, which is shown in Lemma~\ref{Lemma::Proof::5}.
 \begin{lemma}
\label{Lemma::Proof::5}
Given any action sequence $k_1,\dots,k_T$, with the non-adaptive and adaptive exploration rates as in Algorithm~\ref{FTRl::extra_exploration}, we have 
    \begin{align*}
       &\text{Non-Adaptive:} &\E\Big[\sum_{t=1}^T \langle \hat\ell_{t}-\hat\ell_{t}', \p_t'-\p^\dagger \rangle \Big]&\le \ell_{\infty}^-(2n^2+\sqrt{nT})\log_2(1+\ell_{\infty}),\\
       &\text{Adaptive:} &\E\Big[ \sum_{t=1}^T \langle \hat\ell_{t}-\hat\ell_{t}', \p_t'-\p^\dagger \rangle \Big]   &\le \ell_{\infty}^-\Big(2n^2+3\sqrt{n\sum_{t=1}^T \|\ell_t \|_{\infty}}\Big)\log_2(1+\ell_{\infty}).
    \end{align*} 
\end{lemma}

Summing the bounds for $\circled{1}$,$\circled{2}$,$\circled{3}$ gives Theorem~\ref{regret:ALG2}.

\section{Experiments}
We now corroborate our theoretical improvements and testify the performance of our algorithms \texttt{UMAB-G} (Algorithm~\ref{FTRl::general} with non-adaptive exploration) and \texttt{UMAB-G-A} (Algorithm~\ref{FTRl::general} with adaptive exploration).
We compare to \textbf{all} existing scale-free/unbounded AMAB algorithms, including \texttt{SF-MAB} \cite{putta2022scale}, \texttt{SF-MAB-A} \cite{putta2022scale}, \texttt{AHB} \cite{hadiji2020adaptation}, and \texttt{banker-OMD} \cite{huang2023banker}.
The figures show the average performance and standard deviations across $500$ trails.

\begin{figure}[!t]
\centering  
\subfigure[Stock Market Data]{
\label{stock_fig_1}
\includegraphics[width=0.3\textwidth]{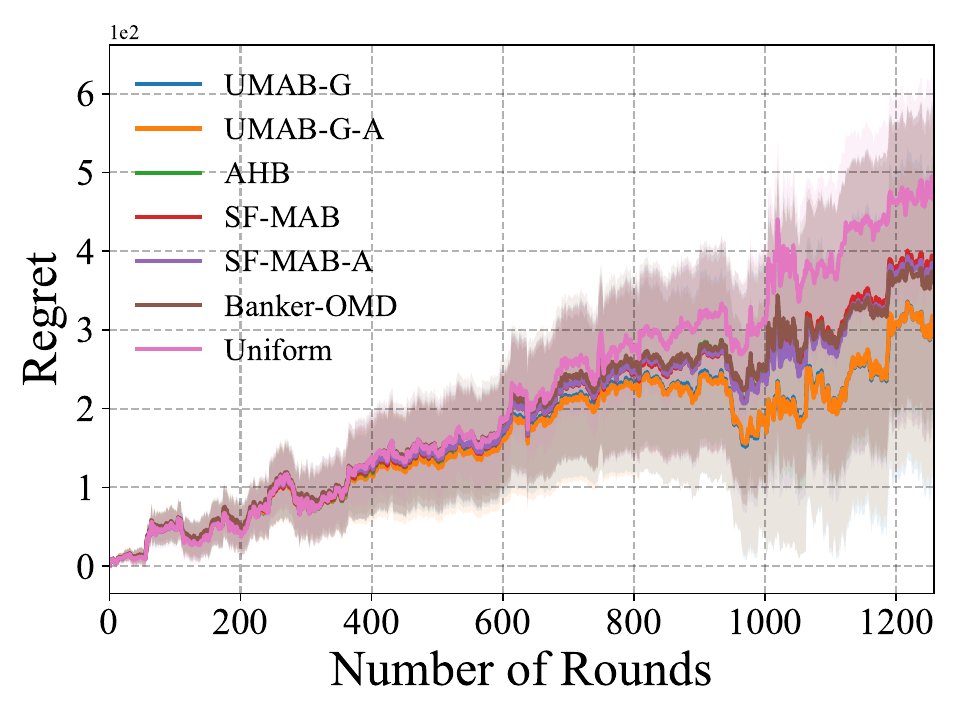}}
\subfigure[Amazon Sales Data]{
\label{amz_fig_1}
\includegraphics[width=0.3\textwidth]{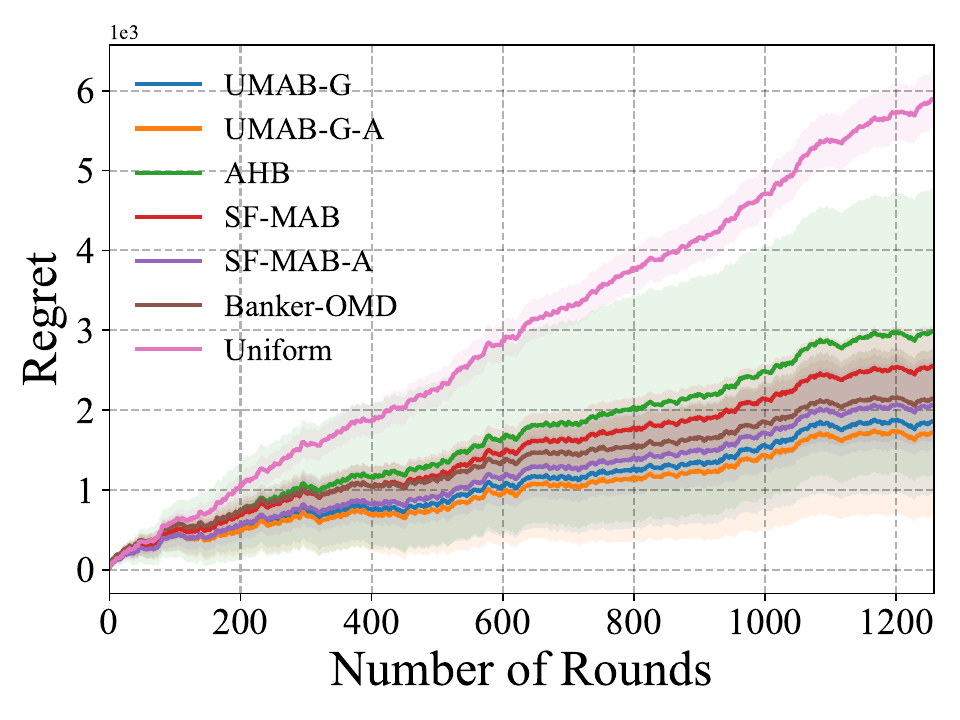}}
\subfigure[Meta Algorithm Selection]{
\label{sgd_fig_1}
\includegraphics[width=0.3\textwidth]{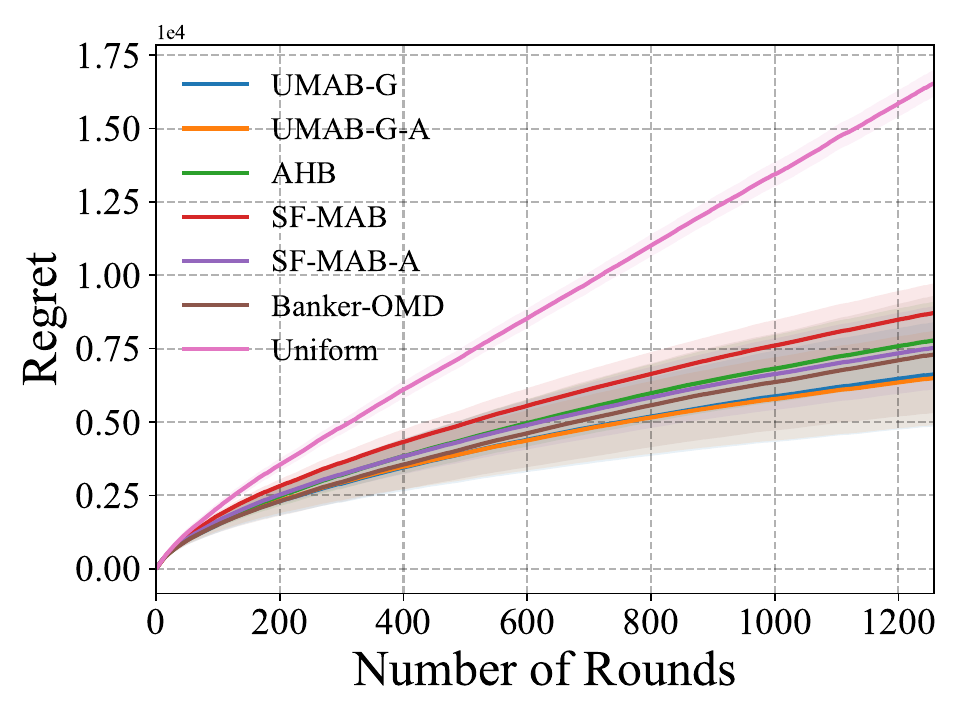}}
\caption{Real Data Experiments}
\label{Fig.1}
\end{figure}

\paragraph{Applications to Stock Trading}: 
In out first experiment, we consider an application to the stock market.
Here we consider $n=10$ stocks and $T=1258$ rounds (daily price for 5-years).
For every stock, its loss is the normalized price difference, i.e., the difference between two consecutive days for $100$ shares. 
Stock prices are generally chaotic and the fluctuation can vary greatly among stocks and across time.
The regret trajectories of the different algorithms are illustrated in Figure~\ref{stock_fig_1}.
Note that the regret of \texttt{UMAB-G} and \texttt{UMAB-G-A} is significantly smaller than that of other algorithms, In particular when the number of rounds is large.
This is because 1). Compared to \cite{putta2022scale}, our algorithms tune the learning and exploration rate more carefully, resulting in a saving of $\mathcal{O}(\sqrt{n})$ term in theory and better empirical performance in practice. 
2). Compared to \cite{huang2023banker}, our exploration rate design ensures that the algorithms can perceive the changes in loss scale and adapt learning rate in time.
3). Compared to \cite{hadiji2020adaptation}, our exploration design leads to smaller regret than mixing with uniform distribution.
 \begin{figure}[!t]
 \centering  
 \subfigure[Extra Exploration Benefits]{
 \label{extra_explo_fig_1}
 \includegraphics[width=0.32\textwidth]{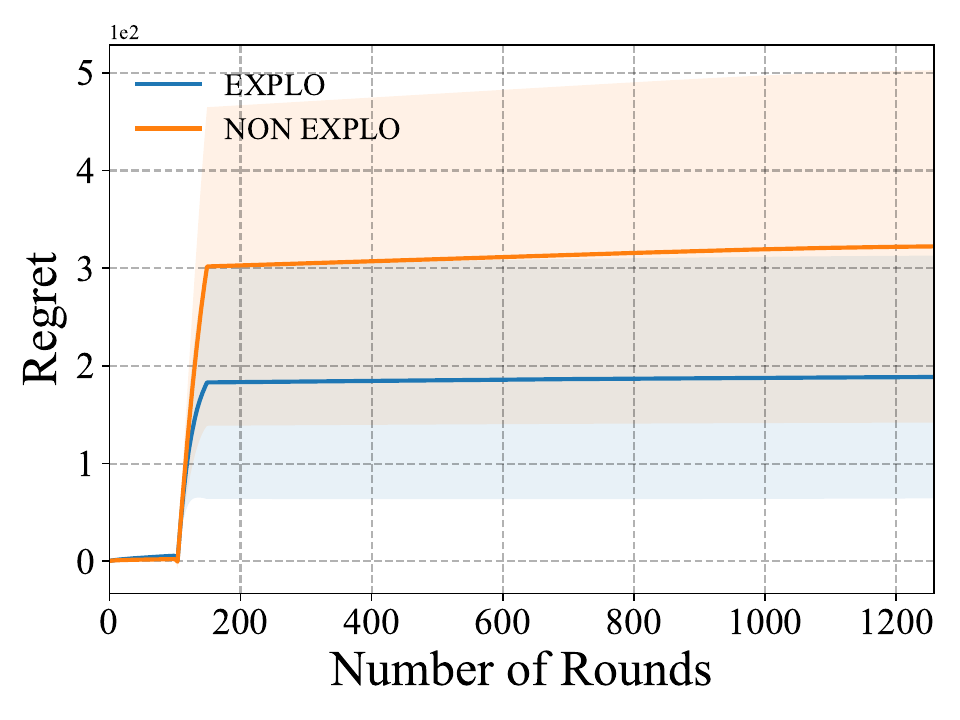}}
 \subfigure[Adaptive Better]{
 \label{adaptive_better_fig_1}
 \includegraphics[width=0.32\textwidth]{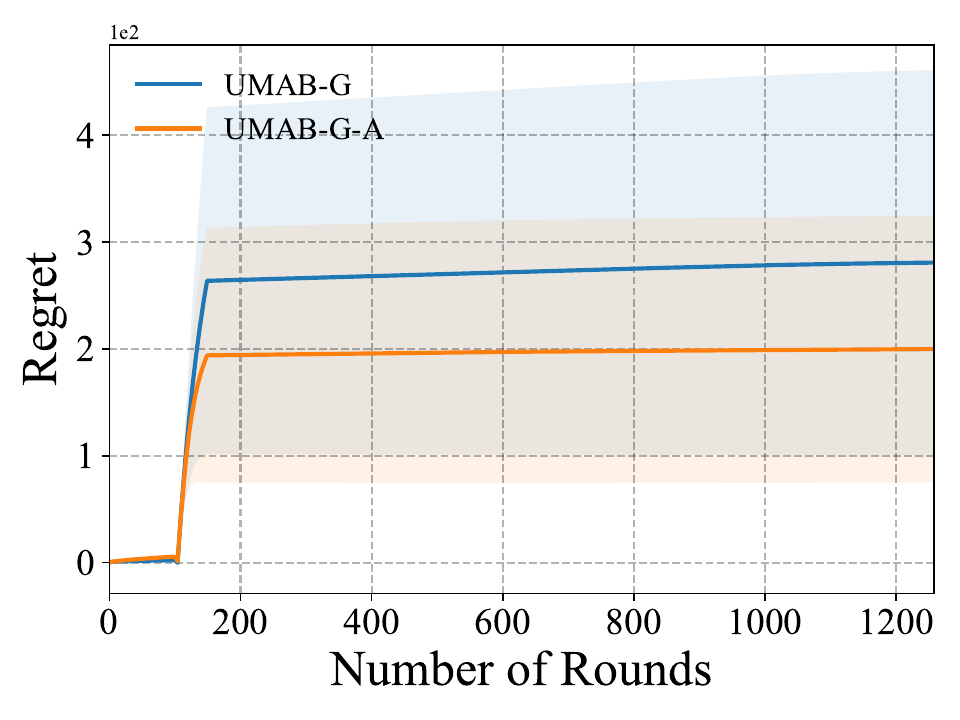}}
 \subfigure[Non-Adaptive Better]{
 \label{non_adaptive_better_fig_1}
 \includegraphics[width=0.32\textwidth]{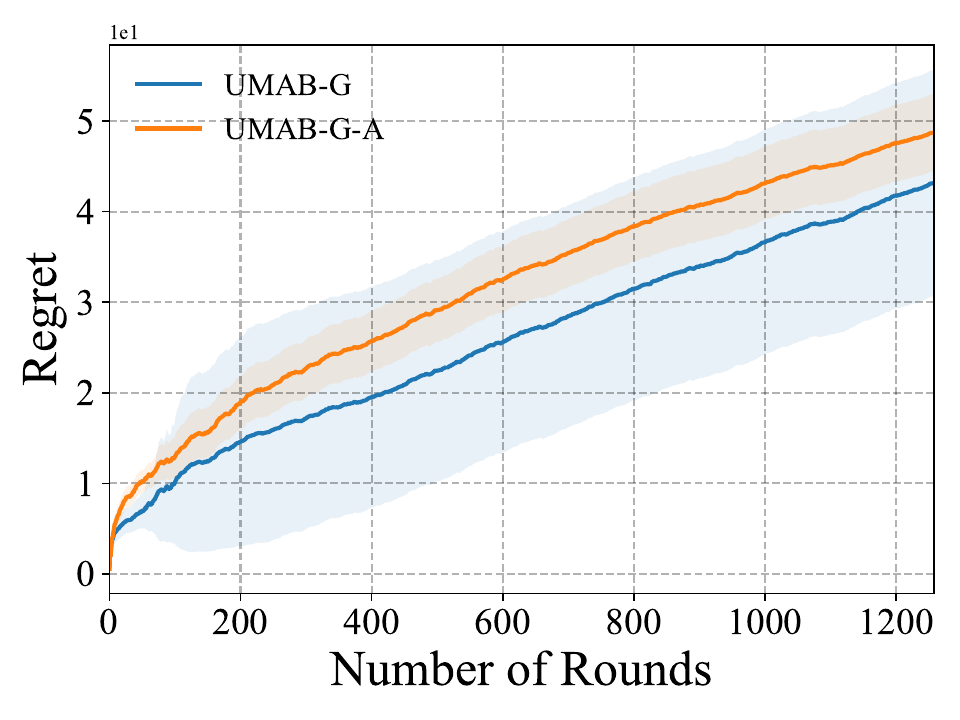}}
 \caption{Impact of Extra Exploration with Non-Adaptive/Adaptive Rates}
 \label{Fig.2}
 \end{figure}

\paragraph{Applications to Amazon Sales}
We further construct an experiment using Amazon sales data.
Similar to the above, we consider $n=10$ Amazon stores and $T=1258$ rounds (weekly sales for 2-years).
We assume that in each round, each store randomly discloses the weekly sales of one of its departments.
The loss is defined by the negative of the weekly sales.
We generate $10$ rounds of loss using one week's data.
Notice that the loss we considered in this setting is completely negative.
The simulation results are shown in Figure~\ref{amz_fig_1}.
As expected, our algorithms outperform all other competitors.
Compared to the stock market example, the fluctuation of regret trajectories of Amazon sales data is more stable for all the algorithms.
This is because changes in Amazon store sales are more gradual than those in stocks: since all the algorithms we consider in the experiment are based on the FTRL/OMD framework, such a loss sequence will induce a stable action distribution, thereby resulting in the smoothness of the regret curve.

\paragraph{Applications to Model Selection}
In the last setting, we explore an application to the model selection problem.
We assume that we have access to $n=10$ linear regression meta-algorithms (SGD with different learning rate).
Similarly to the above, we set the number of rounds $T=1258$.
In each round $t$, the meta-algorithms output the training loss error based on a dataset of size $t$.
Notice that since the size of the data set varies in each round, the optimal meta-algorithm will also change.
In this scenario, the regret measures whether a model selection algorithm can promptly detect the change in the optimal meta-algorithm.
Moreover, the prediction error can be very large when the data set is small.
The results are shown in Figure~\ref{sgd_fig_1}. 
Again, the regrets of our algorithms are strictly smaller than all baselines.
Compared to the first two experiments, the regret trajectories are smoother because of the stochastic nature of the loss sequence as $t$ increases.

\paragraph{Impact of extra exploration}
 We demonstrate the importance of extra exploration for unbounded loss.
Consider a problem with two arms $n=2$ and set $T=1258$.
 We design the following loss sequence:
 \begin{align}
 \label{eq:bad_loss}
\ell_t = 
\begin{cases}
    [0, -0.5]^\top, & \text{if $1\le t<100$}\\
    [-10, 0]^\top, & \text{if $100\le t<150$}\\
    [-0.05, 0]^\top, & \text{if $150\le t<1258$}
\end{cases}
\end{align}
The intuition is to try deceive algorithms into taking the second arm as the ``superior option" in the initial rounds which reduces the frequency of algorithms pulling the first arm, and thus hindering algorithms ability to detect the changes of the optimal arm.
In particular, considering the loss can be unbounded, failing to detect the changes is costly.
 In this case, the regret trajectories are provided in Figure~\ref{extra_explo_fig_1}, where the comparison is between \texttt{UMAB-G-A} and our algorithm with no extra exploration.
 It suffices to note that the algorithm with extra exploration performs much better than the one without extra exploration. 
 This is consistent with the intuition of our design: extra exploration ensures that each arm has a probability of being pulled, so that the algorithm can always perceive the changes in the losses and adjust its learning rate in relatively few rounds.

 \paragraph{Comparison between \texttt{UMAB-G} and \texttt{UMAB-G-A}}
 In the last part we investigate the difference between our algorithms with non-adaptive and adaptive exploration rates.
 Intuitively, adaptive exploration rate is usually larger than the non-adaptive rate because it is of order $O(1/\sqrt{t})$ instead of $O(1/\sqrt{T})$ (assuming $\ell_\infty\ll T$).
 This makes adaptive exploration perform better in adversary cases, e.g. as shown in Figure~\ref{adaptive_better_fig_1}, where we use the same loss sequence in (\ref{eq:bad_loss}).
 However, if the loss sequence is not adversary, e.g. there exists one arm that is always better than the others, non-adaptive exploration will be better since it loses less in extra exploration.
 An example is illustrated in Figure~\ref{non_adaptive_better_fig_1}, where we use stochastic loss with expectation $[1, 0]^\top$.
 In summary, adaptive and non-adaptive have their own advantages under different loss sequences in practice.
\section{Conclusion}
We proposed the first algorithms that achieve optimal adaptive and non-adaptive regrets in adversarial multi-armed bandit problem with unbounded losses. Real data experiments validate our theoretical findings and demonstrate the superior performance of our algorithms compared to all existing algorithms for unbounded losses. Future work include extending our algorithmic tools to more challenging settings such as contextual bandit and reinforcement learning.

\printbibliography
\newpage
\appendix

\section{Additional Discussion about closely related works}
\subsection{Detailed comparison to \cite{putta2022scale}}
In this subsection, we provide a detailed comparison between our work and \cite{putta2022scale} since it is the most closely related work to ours.
Both works are based on FTRL-type algorithms design, and both consider non-adaptive and adaptive extra exploration.
The key idea of \cite{putta2022scale} is to bound (\ref{bound::general}) by $\mathcal{O}(\ell_{t,k_t}^2/p_{t,k})$, resulting in an expectation regret $\mathcal{O}(\|\ell_t\|_2)$.
In our work, we refine the analysis of (\ref{bound::general}), improving the bound to $\mathcal{O}(\ell_{t,k_t}^2)$, where the expectation is bounded by $\mathcal{O}(\|\ell_t\|^2_\infty)$. 
Considering the worst case scenario where $\|\ell_t\|^2_2 = n\|\ell_t\|^2_\infty$, our algorithm saves $\sqrt{n}$ in the regret.

Furthermore, \cite{putta2022scale} choose a uniform distribution for extra exploration. 
This approach ensures an exploration error (\circled{2} in this paper) of $\mathcal{O}(\ell_\infty\sqrt{nT})$ in non-adaptive case. 
However, for adaptive case, mixing a uniform distribution results in a large variance in the analysis of the exploration error.
The proof idea of \cite{putta2022scale} can be summarized as (under our notations definition)
\begin{align*}
    \langle \hat\ell'_{t}, \mathbf{c}_t \rangle\le \|\hat\ell'_{t}\|_\infty\|\mathbf{c}_t\|_1\le \ell_\infty\sqrt{nT}\cdot n = \ell_\infty n^{3/2} \sqrt{T},
\end{align*}
which is suboptimal in $n$.
In this study, we design to a new exploration strategy, as described in Algorithm~\ref{FTRl::extra_exploration}.
By Lemma~\ref{Lemma::Proof::4::2}, we bound $\langle \hat\ell'_{t}, \mathbf{c}_t \rangle$ by $\mathcal{O}(\ell_\infty(\sqrt{nT}+n^2))$, which is optimal in $n$ for large enough $T$.
In summary, the algorithms presented in this article offer a $\mathcal{O}(\sqrt{n})$ improvement of the regret over \cite{putta2022scale}, in both non-adaptive and adaptive settings, thanks to both our novel exploration strategy and tighter analysis.

\subsection{The error in \texttt{Banker-OMD} \cite{huang2023banker}}
\label{Banker_error}
\cite{huang2023banker} shared a similar clipping (skipping) idea with us.
In Lemma 4.2 of \cite{huang2023banker}, the authors control the regret of the general case by the regret of the non-negative case directly (Theorem 4.2 of \cite{huang2023banker}).
In this case, the authors bounded the clipping error (i.e., \circled{3} in this paper) by
\begin{align*}
     \langle \hat\ell_{t}, \p_t-\p^\dagger \rangle\le \langle \hat\ell_{t}, \p_t \rangle \le \ell_\infty.
\end{align*}
However, notice that the above only holds true if $\hat\ell_{t}\ge 0$. 
When $\hat\ell_{t}<0$, -$\langle \hat\ell_{t}, \p^\dagger \rangle$ is positive and on the order of $1/p_{t,k_t}$, which can be arbitrarily unbounded. 
In this case, their regret will always include a $\mathcal{O}(1/p_{t,k_t})$ term and thus be unbounded. We have confirmed this with the authors of \cite{huang2023banker}, and indeed they have made the mistake in their proof. So their current analysis for the general loss setting does not work.

One may think that the issue can be solved by analyzing the regret using $\ell_t$ instead of $\hat \ell_{t}$, i.e.,
\begin{align*}
    \E[  \langle \ell_t, \p_t-\p^\dagger \rangle] = \E[ \mathbbm{1}_{\neg {clip}} \langle \ell_t, \p_t-\p^\dagger \rangle]+\E[ \mathbbm{1}_{\neg clip} \langle \ell_t, \p_t-\p^\dagger \rangle]
\end{align*}
where $\mathbbm{1}_{\neg clip}$ denotes the probability of the clipping happening.
Using the proof of \cite{huang2023banker}, it suffices to show the second term can be bounded by $\mathcal{O}(\ell_\infty\log \ell_\infty)$.
It might be intuitive to think that the first term can also be bounded by using $\hat\ell'$ to estimate $\mathbbm{1}_{\neg clip (t,k)}\ell_t$.
However, we note that
\begin{align*}
    \E[\langle \hat\ell_t',  \p_t-\p^\dagger  \rangle] &= \sum_{k} {p_{t,k}}\Big(\frac{\mathbbm{1}_{\neg clip (t,k)}\ell_{t,k}}{p_{t,k}} p_{t,k} - \frac{\mathbbm{1}_{\neg clip (t,k)}\ell_{t,k}}{p_{t,k}} \mathbbm{1}(k=k^*) \Big)\\
    &= \sum_{k} \mathbbm{1}_{\neg clip (t,k)}\ell_{t,k} x_{t,k} - \sum_{k}\mathbbm{1}_{\neg clip (t,k)}\ell_{t,k}\mathbbm{1}(k=k^*)\\
    &=  \sum_{k} \mathbbm{1}_{\neg clip (t,k)}\ell_{t,k} x_{t,k}- \mathbbm{1}_{\neg clip(t,k^\star)}\ell_{t,k^\star}\\
    &\not= \sum_{k}  x_{t,k} \Big(\mathbbm{1}_{\neg clip (t,k)}(\ell_{t,k}- \ell_{t,k^\star})\Big)\\
    &=  \E[ \mathbbm{1}_{clip(t)} \langle \ell_t, x_t  - y \rangle],
\end{align*}
which implies that $\hat\ell_t'$ is not an unbiased estimator of $\mathbbm{1}_{\neg clip (t,k)}\ell_t$, so this route does not work. Therefore, as far as we can see, there doesn't exist a clear way of fixing the proof in \cite{huang2023banker} to make their results match ours.

In our paper, we avoid issue by adding extra exploration to upper bound $\|\hat \ell_t\|_\infty$.
We suspect such explicit exploration is inevitable for no-regret learning under the unbounded losses \cite{bubeck2012regret}.
Besides this issue, our differences and improvements compared to \cite{huang2023banker} mainly include:
(1). Our results reveal an asymmetry between positive and negative losses in the AMAB problem. 
In particular, there is no clipping in our algorithm \texttt{UMAB-NN}, which greatly simplifies the algorithms in \cite{huang2023banker}.
(2). The space complicity of our algorithms is $\mathcal{O}(n)$ because the algorithm only needs to maintain a constant number of $\R^n$ vectors. In contrast, the space complexity of \cite{huang2023banker} is $\mathcal{O}(T^2)$ due to the necessity of keeping a weight matrix of size $T\times T$.

\section{Proof of Theorem~\ref{regret::ALG1}}
\subsection{Main proof and statement of technical lemmas}
Recall (\ref{Definition::Expected_Regret}), the expected regret is denoted by
\begin{align*}
    \E\Big[   \sum_{t=1}^T \ell_{t,k_t} -  \min_{k\in [n]}  \sum_{t=1}^T \ell_{t,k}\Big] = \E\Big[ \sum_{t=1}^T \langle \ell_t, \p_t-\p^\star \rangle  \Big] = \E\Big[ \sum_{t=1}^T \langle \hat \ell_t, \p_t-\p^\dagger \rangle  \Big]+ \sum_{t=1}^T  \langle \ell_t, \p^\dagger-\p^\star \rangle,
\end{align*}
where $\p^\star$ denote the best fixed strategy.
In particular, we consider
\begin{align*}
    \p^\dagger = \Big(1-\frac{1}{T}\Big)\p^\star+\frac{\mathbf{1}_n}{nT}.
\end{align*}
where $\mathbf{1}_n$ is the all-ones vector. It is obvious that $\p^\dagger\in \Delta_n$. In this case, there is
\begin{align*}
    \langle \ell_t, \p^\dagger-\p^\star \rangle\le  \langle \ell_t, \frac{\mathbf{1}_n}{nT}-\frac{1}{T}\p^\star \rangle\le \frac{1}{nT} \langle \ell_t, \mathbf{1}_n \rangle\le \frac{1}{nT}\|\ell_t\|_1\le \frac{\ell_{\infty}}{T},
\end{align*}
where the second inequality is due to $\ell_t\ge 0$ by assumption.
Thus we have $\sum_{t=1}^T  \langle \ell_t, \p^\dagger-\p^\star \rangle\le \ell_{\infty}$.
It suffices to focus on $ \sum_{t=1}^T \langle \ell_t, \p_t-\p^\star \rangle$.
Recall (\ref{bound3}), there is
\begin{align*}
        \sum_{t=1}^T \langle \hat \ell_t, \p_t-\p^\dagger \rangle&\le \frac{\Psi(\p^\dagger) }{\eta_{T+1}}+ \sum_{t=1}^T \min \Big(     \frac{1}{2}\eta_{t} \ell_{t,k_t}^2, |\ell_{t,k_t}| \Big)\\
       &\le \frac{n\log(nT)}{\eta_{T+1}}+ \sum_{t=1}^T \min \Big(\frac{1}{2}\eta_{t} \ell_{t,k_t}^2, |\ell_{t,k_t}| \Big).
\end{align*}
where the second inequality is because all entries of $\p^\dagger$ are no less than $1/nT$ by definition.

It remains to bound $\sum_{t=1}^T \min \Big(\frac{1}{2}\eta_{t} \ell_{t,k_t}^2, |\ell_{t,k_t}| \Big)$.
The proof relies on a technical lemma from \cite{orabona2018scale}.
\begin{lemma}(\cite{orabona2018scale} Lemma 3)
    \label{Theorem::Proof::1::1}
    Let $a_1,\dots,a_T\ge 0$. Then
    \begin{align*}
        \sum_{t=1}^T \min\Big(  \frac{a_t^2}{  \sqrt{\sum_{s=1}^{t-1} a_s^2  }}, a_t  \Big)\le 3.5 \sqrt{\sum_{t=1}^{T} a_t^2} +3.5\max_{t\in [T]} a_t
    \end{align*}
\end{lemma}
Using Lemma~\ref{Theorem::Proof::1::1} and $\eta_0,\dots,\eta_{T}$ as in Algorithm~\ref{FTRl::non-negative}, we have
\begin{align*}
    \sum_{t=1}^T \langle \hat \ell_t, \p_t-\p^\dagger \rangle&\le \frac{n\log(nT)}{\eta_{T+1}}+ \sum_{t=1}^T \min \Big(\frac{1}{2}\eta_{t} \ell_{t,k_t}^2, |\ell_{t,k_t}| \Big)\\
    &\le \frac{1}{2}\sqrt{n\sum_{t=1}^T \ell_{t,k_t}^2 }\log(nT) + \sum_{t=1}^T \min \Big(\sqrt{\frac{n}{\sum_{s=1}^{t-1} \ell_{s,k_s}^2}}\ell_{t,k_t}^2, |\ell_{t,k_t}| \Big)\\
    &\le \frac{1}{2}\sqrt{n\sum_{t=1}^T \ell_{t,k_t}^2 }\log(nT)+ \sqrt{n} \sum_{t=1}^T \min \Big(\frac{\ell_{s,k_t}^2}{ \sqrt{ \sum_{s=1}^{t-1} \ell_{s,k_s}^2}}, |\ell_{t,k_t}| \Big)\\
    &\le \frac{1}{2}\sqrt{n\sum_{t=1}^T \ell_{t,k_t}^2 }\log(nT) + 3.5\sqrt{n} \Big( \sqrt{\sum_{t=1}^T \ell_{t,k_t}^2 } + \max_{t\in [T]} |\ell_{t,k_t}| \Big)\\
    &\le 4\sqrt{n\sum_{t=1}^T \ell_{t,k_t}^2 }\log(nT)+3.5\sqrt{n}\ell_{\infty}\\
    &\le 4\sqrt{n\sum_{t=1}^T \|\ell_t\|_\infty^2 }\log(nT)+3.5\sqrt{n}\ell_{\infty}.
\end{align*}
Note that the right hand side of the above is deterministic.
Thus
\begin{align*}
    \E\Big[   \sum_{t=1}^T \ell_{t,k_t} -  \min_{k\in [n]}  \sum_{t=1}^T \ell_{t,k}\Big] &\le 4\sqrt{n\sum_{t=1}^T \|\ell_t\|_\infty^2 }\log(nT)+3.5\sqrt{n}\ell_{\infty}+\ell_{\infty}\\
    &\le \tilde{\mathcal{O}}\Big(\sqrt{n\sum_{t=1}^T \|\ell_t\|_\infty^2 } \Big)
\end{align*}
completes the proof.

\subsection{Proof of technical Lemmas}
\subsubsection{Proof of Lemma~\ref{Lemma::FTRL_regret}}
For notations simplicity, we denote by 
\begin{align*}
    \Psi_{t}(\p) = \frac{1}{\eta_{t}}\Psi(\p).
\end{align*}
We first note
\begin{align*}
    \sum_{t=1}^T \langle \hat \ell_t, \p_t-\p^\dagger \rangle &=   -F_{T+1}(\p^\dagger)+\Psi_{T+1}(\p^\dagger)+\sum_{t=1}^T \langle \hat \ell_t, \p_t \rangle\\
    &=-F_{T+1}(\p^\dagger)+\Psi_{T+1}(\p^\dagger)-F_1(\p_1)+F_{T+1}(\p_{T+1}) \nonumber \\  
    &+\sum_{t=1}^T \left( F_t(\p_t)-F_{t+1}(\p_{t+1}) \right)+\sum_{t=1}^T \langle \hat \ell_t, \p_t \rangle\\
    &=-F_{T+1}(\p^\dagger)+\Psi_{T+1}(\p^\dagger)-F_1(\p_1)+F_{T+1}(\p_{T+1}) \nonumber \\  
    &+\sum_{t=1}^T \left( F_t(\p_t)+\langle \hat \ell_t, \p_t \rangle-F_{t+1}(\p_{t+1}) \right)
\end{align*}
By definition, there is 
\begin{align*}
    F_{T+1}(\p_{T+1})-F_{T+1}(\p^\dagger)&=\min_{\p\in \Delta_n}F_{T+1}(\p)-F_{T+1}(\p^\dagger) \le 0\\
    \Psi_{T+1}(\p^\dagger)-F_1(\p_1) &= \Psi_{T+1}(\p^\dagger)- \min_{\p\in \Delta_n }\Psi_{1}(\p)\le  \Psi_{T+1}(\p^\dagger).
\end{align*}
Thus, we obtain 
\begin{align*}
    \sum_{t=1}^T \langle \hat \ell_t, \p_t-\p^\dagger \rangle\le \Psi_{T+1}(\p^\dagger) + \sum_{t=1}^T \left( F_t(\p_t)+\langle \hat \ell_t, \p_t \rangle-F_{t+1}(\p_{t+1}) \right)
\end{align*}
Furthermore, we note that
\begin{align*}
     F_t(\p_t)+\langle \hat \ell_t, \p_t \rangle-F_{t+1}(\p_{t+1})&=\sum_{s=1}^{t}\langle \hat \ell_s, \p_t-\p_{t+1} \rangle + \frac{1}{\eta_{t}}\Psi(\p_t)- \frac{1}{\eta_{t+1}}\Psi(\p_t) \\
     &\le \sum_{s=1}^{t}\langle \hat \ell_s, \p_t-\p_{t+1} \rangle + \frac{1}{\eta_{t}}\Psi(\p_t)- \frac{1}{\eta_{t}}\Psi(\p_t)\\
     &=\langle \hat \ell_t, \p_t-\p_{t+1} \rangle+ F_t(\p_t)-F_{t}(\p_{t+1}),
\end{align*}
where the first inequality is due to the assumption $\eta_{t+1}\le \eta_{t}$.
Combining the above concludes the proof.

\subsubsection{Proof of Lemma~\ref{Lemma::local_norm}}
We first prove inequality (\ref{bound::general}).
By Taylor’s expansion, 
\begin{align*}
    F_t(\p_{t+1})-F_t(\p_t) = \langle \nabla F_t(\p_t), \p_{t+1}-\p_t \rangle+\frac{1}{2}\|\p_{t+1}-\p_t\|^2_{\nabla^2 F_t(\xi_t)}.
\end{align*}
where $\xi_t=\alpha \p_t+(1-\alpha)\p_{t+1}$ for some $\alpha\in [0,1]$.
By definition,
\begin{align*}
    \p_t =\arg\min_{\p\in \Delta_n} F_t(\p).
\end{align*}
By KKT conditions, there exists some $\lambda_t\in \R$ such that
\begin{align*}
     \p_t =\arg\min_{\p\in \R} \Big(F_t(\p) +\lambda_t (1-\sum_{k=1}^n p_{t,k})\Big).
\end{align*}
By the optimality of $\p_t$, we have
\begin{align*}
    \nabla F_t(\p_t) +\lambda_t\mathbf{1}_n = 0,
\end{align*}
which implies 
\begin{align*}
    \langle \nabla F_t(\p), \p_{t+1}-\p_t \rangle =  \langle -\lambda_t\mathbf{1}_n, \p_{t+1}-\p_t \rangle = 0. 
\end{align*}
Thus, there is 
\begin{align*}
    F_t(\p_{t+1})-F_t(\p_t) &= \frac{1}{2}\|\p_{t+1}-\p_t\|^2_{\nabla^2 F_t(\xi_t)}.
\end{align*}
Using the above, 
\begin{align*}
     \langle \hat \ell_t, \p_t-\p_{t+1} \rangle+ F_t(\p_t)-F_{t}(\p_{t+1})&= \langle \hat \ell_t, \p_t-\p_{t+1} \rangle-\frac{1}{2}\|\p_{t+1}-\p_t\|^2_{\nabla^2 F_t(\xi_t)}  \\
    & \le \max_{\p\in \R} \Big( \langle \hat \ell_t, \p \rangle-\frac{1}{2}\|\p\|^2_{\nabla^2 F_t(\xi_t)} \Big) \\
    &\le \frac{1}{2}\|\hat \ell_t\|^2_{(\nabla^2 F_t(\xi_t))^{-1}} = \frac{1}{2}\eta_{t}\|\hat \ell_t\|^2_{(\nabla^2 \Psi(\xi_t))^{-1}},
\end{align*}
where the second inequality is because $\nabla^2 \Psi(\xi_t)$ is a diagonal matrix and the second equality is due to $\nabla^2 F_t(\xi_t) = \nabla^2 \Psi(\xi_t)/\eta_{t} $.
Thus the proof of (\ref{bound::general}) is complete.

Now we prove
\begin{align*}
    \langle \hat \ell_t, \p_t-\p_{t+1} \rangle+ F_t(\p_t)-F_{t}(\p_{t+1})\le \frac{1}{2}\eta_{t}\| \hat \ell_t\|^2_{(\nabla^2 
 \Psi(\p_t))^{-1}}
\end{align*}
if $\hat\ell_t\in \R_+^n$.
Recall
\begin{align*}
     \| \hat \ell_t\|^2_{(\nabla^2  \Psi(\xi_t))^{-1}} = \sum_{k=1}^n \frac{\hat \ell_{t,k}^2 }{\nabla^2_{k,k}  \Psi(\xi_t)}  = \sum_{k=1}^n \frac{ \ell_{t,k}^2 \mathbbm{1}(k=k_t) }{p_{t,k}^2} \xi_{t,k}^2 = \frac{ \ell_{t,k_t}^2 }{p_{t,k_t}^2} \xi_{t,k_t}^2
\end{align*}
and $\xi_t$ is between $\p_t$ and $\p_{t+1}$, we prove case by case.
\begin{enumerate}
    \item ($ p_{t,k_t}- p_{t+1,k_t}<0$): In this case, we have 
    \begin{align*}
        \langle \hat \ell_t, \p_t-\p_{t+1} \rangle+ F_t(\p_t)-F_{t}(\p_{t+1})&\le \langle \hat \ell_t, \p_t-\p_{t+1} \rangle\\
        &= \hat \ell_{t,k_t}(p_{t,k_t}- p_{t+1,k_t})\\
        &\le 0 \le \frac{1}{2} \| \hat \ell_t\|^2_{(\nabla^2  \Psi(\p_t))^{-1}}.
    \end{align*}
    The first inequality is due to $\p_t$ minimizing $F_t$.
    \item ($ p_{t,k_t}- p_{t+1,k_t}\ge 0$): 
    In this case, we have $\xi_{t,k_t}\le p_{t,k_t}$, and thus 
    \begin{align*}
        \| \hat \ell_t\|^2_{(\nabla^2  \Psi(\xi_t))^{-1}}\le \ell_{t,k_t}^2 = \| \hat \ell_t\|^2_{(\nabla^2  \Psi(\p_t))^{-1}}
    \end{align*}
    completes the proof.
\end{enumerate}

\subsubsection{Proof of Lemma~\ref{Theorem::Proof::1::1}}
The proof refers to Lemma 3 in \cite{orabona2018scale}.
Without loss of generality, we can assume $a_t>0$, otherwise we can remove all $a_t=0$ without affecting either side of the inequality.
Let $M_t=\max_{s\in [t]} a_s$ and $M_0=0$.
We aim to prove for any $\alpha>1$
\begin{align*}
     \min\Big(  \frac{a_t^2}{  \sqrt{\sum_{s=1}^{t-1} a_s^2  }}, a_t  \Big)\le 2\sqrt{1+\alpha^2}\Big(  \sqrt{\sum_{s=1}^t a_s^2}-\sqrt{\sum_{s=1}^{t-1} a_s^2}   \Big)+\frac{\alpha}{\alpha-1}(M_t-M_{t-1}).
\end{align*}
from which Lemma~\ref{Theorem::Proof::1::1} follows by summing over $t=1,\dots,T$ and choosing $\alpha=\sqrt{2}$.
The proof is based on case analysis.
\begin{enumerate}
    \item ($a_t^2\le \alpha^2 \sum_{s=1}^{t-1} a_s^2 $)
    \begin{align*}
        \min\Big(  \frac{a_t^2}{  \sqrt{\sum_{s=1}^{t-1} a_s^2  }}, a_t  \Big)&\le \frac{a_t^2}{  \sqrt{\sum_{s=1}^{t-1} a_s^2  }} = \frac{a_t^2}{  \sqrt{ 
 \frac{1}{1+\alpha^2} (\alpha^2\sum_{s=1}^{t-1} a_s^2  +\sum_{s=1}^{t-1} a_s^2  ) }}\\
 &\le \frac{a_t^2(1+\alpha^2)}{  \sqrt{a_t^2 + \sum_{s=1}^{t-1} a_s^2  }}\le 2\sqrt{1+\alpha^2}\Big(  \sqrt{\sum_{s=1}^t a_s^2}-\sqrt{\sum_{s=1}^{t-1} a_s^2}   \Big)
    \end{align*}
where the last inequality is by $x^2/\sqrt{x^2+y^2}\le 2(\sqrt{x^2+y^2}-\sqrt{y^2})$.
\item ($a_t^2> \alpha^2 \sum_{s=1}^{t-1} a_s^2 $)
\begin{align*}
    \min\Big(  \frac{a_t^2}{  \sqrt{\sum_{s=1}^{t-1} a_s^2  }}, a_t  \Big)\le a_t = \frac{\alpha a_t-a_t}{\alpha-1}\le \frac{\alpha }{\alpha-1}\Big(a_t-\sqrt{\sum_{s=1}^{t-1} a_s^2  }\Big) \le \frac{\alpha }{\alpha-1}(M_t-M_{t-1}),
\end{align*}
where we use $a_t=M_t$ and $M_{t-1}\le \sqrt{\sum_{s=1}^{t-1} a_s^2  }$.
\end{enumerate}

\section{Proof of Theorem~\ref{regret:ALG2}}
\subsection{Main proof and statement of technical lemmas}
By Lemma~\ref{Lemma::local_norm}, we have 
\begin{align*}
    \langle \hat \ell_t', \p_t-\p_{t+1} \rangle+ F_t(\p_t)-F_{t}(\p_{t+1}) \le \min \Big(     \frac{1}{2}\eta_{t} {\ell'}_{t,k_t}^2, |\ell'_{t,k_t}| \Big).
\end{align*}
if $\ell_{t,k_t}\ge 0$.
Alternatively, when $\ell_{t,k_t}< 0$, by Lemma~\ref{Lemma::local_norm} and \ref{Lemma::Proof::3} and inequality (\ref{bound4}), we have
\begin{align*}
    \langle \hat \ell_t', \p_t-\p_{t+1} \rangle+ F_t(\p_t)-F_{t}(\p_{t+1})&\le \frac{1}{2}\eta_{t}\| \hat \ell_t'\|^2_{(\nabla^2 
 \Psi(\xi_t))^{-1}} = \frac{1}{2}\eta_{t} \frac{ {\ell'}_{t,k_t}^2 }{
{p'}_{t,k_t}^2} \xi_{t,k_t}^2\\
 &\le \frac{1}{2}\eta_{t} {\ell'}_{t,k_t}^2 \frac{ {p}_{t,k_t}^2   }{   {p'}_{t,k_t}^2  } \frac{\max(p_{t,k_t}^2, p_{t+1,k_t}^2)}{{p}_{t,k_t}^2}\\
 &\le 72\eta_{t} {\ell'}_{t,k_t}^2.
\end{align*}
Moreover, we further note by Lemma~\ref{Lemma::Proof::3},
\begin{align*}
    \langle \hat \ell_t', \p_t-\p_{t+1} \rangle+ F_t(\p_t)-F_{t}(\p_{t+1})&\le  \langle \hat \ell_t', \p_t-\p_{t+1} \rangle\\
    & \le \Big| \frac{{\ell'}_{t,k_t}}{{p}_{t,k_t}}\Big| \Big|\frac{ {p}_{t,k_t}   }{   {p'}_{t,k_t}  }\Big| |p_{t,k_t}- p_{t+1,k_t}|\\
    &\le \Big| \frac{{\ell'}_{t,k_t}}{{p}_{t,k_t}}\Big| \Big|\frac{ {p}_{t,k_t}   }{   {p'}_{t,k_t}  }\Big| |5 p_{t,k_t}| \le 10|{\ell'}_{t,k_t}|.
\end{align*}
Combining the above we have 
\begin{align*}
    \langle \hat \ell_t', \p_t-\p_{t+1} \rangle+ F_t(\p_t)-F_{t}(\p_{t+1})\le 18\min \Big(     4\eta_{t} {\ell'}_{t,k_t}^2, |\ell'_{t,k_t}| \Big)
\end{align*}
for any $\ell_t\in \R^n$.
Using a similar proof as in Theorem~\ref{regret::ALG1}, we have
\begin{align*}
    \sum_{t=1}^T \langle \hat \ell_t', \p_t-\p^\dagger \rangle&\le \frac{n\log(nT)}{\eta_{T+1}}+ 18\min \Big( 4\eta_{t} {\ell'}_{t,k_t}^2, |\ell'_{t,k}| \Big)\\
    &\le 4\sqrt{2n^2 \ell^2_{\infty}+n\sum_{t=1}^T {\ell'}_{t,k_t}^2 }\log(nT) + 18\sqrt{n}\min \Big(   \frac{{\ell'}_{t,k_t}^2}{\sqrt{\sum_{s=1}^{t-1} {\ell'}_{s,k_s}^2} }, |\ell'_{t,k}| \Big)\\
    &\le 4\sqrt{2n^2 \ell^2_{\infty}+n\sum_{t=1}^T {\ell'}_{t,k_t}^2 }\log(nT) + 63\sqrt{n} \Big( \sqrt{\sum_{t=1}^T {\ell'}_{t,k_t}^2 } + \max_{t\in [T]} |\ell'_{t,k}| \Big)\\
    &\le 67 \sqrt{2n^2 \ell^2_{\infty}+ n\sum_{t=1}^T {\ell'}_{t,k_t}^2 }\log(nT)+63\sqrt{n}\max_{t\in [T]} |\ell'_{t,k}|\\
    &\le 67\sqrt{2n^2 \ell^2_{\infty}+n\sum_{t=1}^T \|\ell_t\|_\infty^2 }\log(nT)+63\sqrt{n}\ell_{\infty}.
\end{align*}
The last inequality is because $|{\ell'}_{t,k_t}|\le |{\ell}_{t,k_t}|$.
In short, we can bound
\begin{align}
\label{bound6}
    \E\Big[\sum_{t=1}^T \langle \hat \ell_t', \p_t-\p^\dagger \rangle\Big]\le \tilde{\mathcal{O}}\Big( \sqrt{n^2 \ell_{\infty}^2+n\sum_{t=1}^T \|\ell_t\|_\infty^2 }  \Big).
\end{align}

Now we summarize all the results.
\begin{align*}
     &\E\Big[   \sum_{t=1}^T \ell_{t,k_t} -  \min_{k\in [n]}  \sum_{t=1}^T \ell_{t,k}\Big]\\
     =& \E\Big[   \sum_{t=1}^T \langle \hat\ell_{t}, \p_t'-\p^\star \rangle \Big]\\
     \le& \E\Big[   \sum_{t=1}^T \langle \hat\ell_{t}, \p_t'-\p^\dagger \rangle \Big]+\ell_\infty\\
     \le& \E\Big[{\sum_{t=1}^T \langle \hat\ell_{t}', \p_t-\p^\dagger \rangle}\Big]+
    \E\Big[{\sum_{t=1}^T \langle \hat\ell'_{t}, \p_t'-\p_t \rangle}\Big]+
    \E\Big[{\sum_{t=1}^T \langle \hat\ell_{t}-\hat\ell_{t}', \p_t'-\p^\dagger \rangle}\Big]+\ell_\infty.
\end{align*}
Based on Lemma~\ref{Lemma::Proof::4} and \ref{Lemma::Proof::5} and inequality (\ref{bound6}), we have
\begin{enumerate}
    \item (Non-Adaptive):
    \begin{align*}
        &\E\Big[   \sum_{t=1}^T \ell_{t,k_t} -  \min_{k\in [n]}  \sum_{t=1}^T \ell_{t,k}\Big]\\
        \le& \tilde{\mathcal{O}}\Big( \sqrt{n^2\ell_{\infty}^2+n\sum_{t=1}^T \|\ell_t\|_\infty^2 }  \Big)+\tilde{\mathcal{O}}\Big(  \sqrt{n\sum_{t=1}^T \|\ell_t\|^2_\infty} \Big)+\tilde{\mathcal{O}}\Big( \ell_\infty^{-} (n^2+\sqrt{nT}) \Big) \\
        =&  \tilde{\mathcal{O}}\Big( \ell_\infty n^2+\sqrt{n\sum_{t=1}^T \|\ell_t\|^2_\infty}+\ell_\infty^{-} \sqrt{nT} \Big).
    \end{align*}
    \item (Adaptive):
    \begin{align*}
        &\E\Big[   \sum_{t=1}^T \ell_{t,k_t} -  \min_{k\in [n]}  \sum_{t=1}^T \ell_{t,k}\Big]\\
        \le& \tilde{\mathcal{O}}\Big( \sqrt{n^2\ell_{\infty}^2+n\sum_{t=1}^T \|\ell_t\|_\infty^2 }  \Big)+\tilde{\mathcal{O}}\Big( \ell_\infty \Big(n^2+\sqrt{n \sum_{t=1}^{T} \|\ell_t \|_{\infty}  } \Big) +\sqrt{n \sum_{t=1}^{T} \|\ell_t \|_{\infty}}  \Big)\\
        &+\tilde{\mathcal{O}}\Big( \ell_\infty^- \Big(n^2+\sqrt{n \sum_{t=1}^{T} \|\ell_t \|_{\infty}  }\Big) \Big) \\
        =&  \tilde{\mathcal{O}}\Big( \ell_\infty n^2+ \sqrt{n\sum_{t=1}^T \|\ell_t\|_\infty^2 } +\ell_\infty \sqrt{n\sum_{t=1}^{T} \|\ell_t \|_{\infty}} + \sqrt{n \sum_{t=1}^{T} \|\ell_t \|_{\infty}} \Big).
    \end{align*}
\end{enumerate}

\subsection{Proof of technical Lemmas}
\subsubsection{Proof of Lemma~\ref{Lemma::Proof::3}}
The first inequality $p_{t,k_t} \le 2p_{t,k_t}'$ can be easily verified.
Recall $\p_t' = \p_t+\rho_t \mathbf{c}_t$ and $k_t^\star \in  \arg\max_{k'\in [n]} p_{t,k'}$ as in Algorithm~\ref{FTRl::extra_exploration}, it suffices to focus on the case $ k_t=k_t^\star$, otherwise $p_{t,k_t}\le p_{t,k_t}'$.
When $ k_t=k_t^\star$, we note that
\begin{align*}
    p_{t,k_t}' = p_{t,k_t}+{\rho_t}c_{t,k_t} \ge  p_{t,k_t} - \frac{1}{2n^2}n = p_{t,k_t}-\frac{1}{2n}.
\end{align*}
The first inequality is due to $\rho_t\le 1/2n^2$ and $c_{t,k_t}\ge -n$ by definition.
Moreover, there is 
\begin{align*}
    p_{t,k_t} \in  \arg\max_{k'\in [n]}p_{t,k'}\ge \frac{1}{n}.
\end{align*}
Thus
\begin{align*}
    p_{t,k_t}\le  p_{t,k_t}+ p_{t,k_t}-\frac{1}{n} = 2 \Big(p_{t,k_t}-\frac{1}{2n} \Big) = 2 p_{t,k_t}'
\end{align*}
completes the proof.

The proof of the second inequality relies on the following two technical lemmas.
\begin{lemma}
\label{Lemma::Proof::3::1}
Given any $L\in \R^n$ and $k\in [n]$, consider
\begin{align*}
        \mathbf{x} &= \arg\min_{\p\in \Delta_n}\Big( \langle L, \p\rangle+\frac{1}{\eta}\Psi(\p)   \Big)\\
        \tilde{\mathbf{x}} &= \arg\min_{\p\in \Delta_n}\Big(  \langle L+\frac{l}{x_k}\mathbf{e}_k, \p\rangle+ \frac{1}{\eta}\Psi(\p)   \Big)
\end{align*}
where $x_k$ is the $k$th entry of $\mathbf{x}$.
If \begin{align*}
   -\frac{1}{2\eta} \le  l\le 0,
\end{align*}
then 
\begin{align*}
    \tilde{x}_k\le 2 x_k.
\end{align*}
\end{lemma}

\begin{lemma}
\label{Lemma::Proof::3::2}
    Given any $L\in \R^n$, consider
\begin{align*}
        \mathbf{x} &= \arg\min_{\p\in \Delta_n}\Big( \langle L, \p\rangle+\frac{1}{\eta}\Psi(\p)   \Big)\\
        \mathbf{x}^\prime &= \arg\min_{\p\in \Delta_n}\Big(  \langle L, \p\rangle+\frac{1}{\eta'} \Psi(\p)   \Big),
\end{align*}
if
\begin{align*}
   \eta^\prime \le \eta\le  C\eta^\prime,
\end{align*}
for some $C>0$, then 
\begin{align*}
    x_k^\prime\le C x_k,\ \forall k\in [n].
\end{align*}
\end{lemma}

Now we use Lemma~\ref{Lemma::Proof::3::1} and \ref{Lemma::Proof::3::2} to bound the magnitude of $p_{t+1,k_t}/p_{t,k_t}$.
Recall the update rule of action distribution
\begin{align*}
    \p_t &= \arg\min_{\p\in \Delta_n}\Big( \langle \sum_{s=1}^{t-1} \hat \ell_s', \p\rangle+ 
\frac{1}{\eta_{t}} \Psi(\p)   \Big),\\
    \p_{t+1} &= \arg\min_{\p\in \Delta_n}\Big( \langle \hat \ell_t'+ \sum_{s=1}^{t-1} \hat \ell_s', \p\rangle+\frac{1}{\eta_{t+1}}\Psi(\p)   \Big).
\end{align*}
Define the intermediate distribution
\begin{align*}
    \tilde{\p}_t &= \arg\min_{\p\in \Delta_n}\Big( \langle \hat \ell_t'+ \sum_{s=1}^{t-1} \hat \ell_s', \p\rangle+\frac{1}{\eta_{t}}\Psi(\p)   \Big).
\end{align*}
Notice that $\hat \ell_t' = \ell_{t,k_t}'\mathbf{1}_{k_t}/p_{t,k_t}$.
Denote by $L=\sum_{s=1}^{t-1} \hat \ell_s'$, by Lemma~\ref{Lemma::Proof::3::1}, ${\tilde{p}_{t,k_t}}/{p_{t,k_t}}\le 2$ if $-{1}/{2\eta_{t}} \le  \ell_{t,k_t}' \le 0$.
Moreover, by Lemma~\ref{Lemma::Proof::3::2}, ${p_{t+1,k_t}}/{p_{t,k_t}}\le 3$ if $\eta_{t+1}\le \eta_{t}\le 3\eta_{t+1}$.
Combining these two results leads to ${p_{t+1,k_t}}/{p_{t,k_t}}\le 6$, which completes the proof.
Therefore, it remains to show that the two conditions hold.

We first prove $-{1}/{2\eta_{t}} \le  \ell_{t,k_t}' \le 0$.
Recall
\begin{align*}
        \eta_{t} = \frac{1}{4}\sqrt{\frac{n}{nC^2_{t}+\sum_{s=1}^{t-1} {\ell^\prime}_{s,k_s}^2}}.
\end{align*}
We have 
\begin{align*}
    {\ell'}_{t,k_t}^2\le 4C_t^2\le 4\Big( \frac{nC_{t}^2+\sum_{s=1}^{t-1} {\ell^\prime}_{s,k_s}^2}{n} \Big)\le \frac{1}{4\eta_{t}^2},
\end{align*}
where the first inequality is by the assumption $\ell_{t,k_t}\le 0$, which implies $\ell_{t,k_t}'\le 0$, and the clipping rule (line 5 of Algorithm~\ref{FTRl::general}).

Then we show $\eta_{t+1}\le \eta_{t}\le 3\eta_{t+1}$.
Since $\eta_{t+1}\le \eta_{t}$ is trivial, it suffices to prove $\eta_{t}\le 3\eta_{t+1}$.
Notice that
\begin{align*}
    C_{t+1}^2 = \min\Big( C_t^2, {\ell'}_{t,k_t}^2  \Big)\le \min\Big( C_t^2, 4 C_t^2 \Big) = 4C_t^2.
\end{align*}
Thus, 
\begin{align*}
    \eta_{t} &= \frac{1}{4}\sqrt{\frac{n}{nC^2_{t}+\sum_{s=1}^{t-1} {\ell^\prime}_{s,k_s}^2}}\\
    &= \frac{3}{4}\sqrt{\frac{n}{9nC^2_{t}+9\sum_{s=1}^{t-1} {\ell^\prime}_{s,k_s}^2}}\\
    &\le \frac{3}{4}\sqrt{\frac{n}{4nC^2_{t}+4nC^2_{t}+\sum_{s=1}^{t-1} {\ell^\prime}_{s,k_s}^2}}\\
    &\le \frac{3}{4}\sqrt{\frac{n}{nC_{t+1}^2+{\ell'}_{t,k_t}^2+\sum_{s=1}^{t-1} {\ell^\prime}_{s,k_s}^2}}\\
    & = 3\eta_{t+1}.
\end{align*}
completes the proof.

\subsubsection{Proof of Lemma~\ref{Lemma::Proof::4}}
\textbf{Non-Adaptive exploration}:
\begin{align*}
     \E\Big[\sum_{t=1}^T \langle \hat\ell'_{t}, \p_t'-\p_t \rangle\Big] &= \E\Big[\sum_{t=1}^T {\rho_t}\langle \hat \ell'_{t}, \mathbf{c}_t \rangle \Big]\\
     &\le \E\Big[\sum_{t=1}^T {\rho_t}\langle |\hat \ell_{t}|, |\mathbf{c}_t| \rangle \Big]\\
     &= \sum_{t=1}^T {\rho_t}\langle |\ell_{t}|, |\mathbf{c}_t| \rangle \\
     &\le 2n\sum_{t=1}^T \frac{\|\ell_t\|_{\infty}}{n^2+\sqrt{nT}}\\
     &\le 2\sqrt{n}  \frac{ \sum_{t=1}^T \|\ell_t\|_{\infty}}{\sqrt{T}}\\
     &\le 2\sqrt{n \sum_{t=1}^T \|\ell_t\|^2_{\infty} }.
\end{align*}
The first inequality is due to that $\hat\ell_t'$ is the truncation of $\hat\ell_t$, thus $|\hat\ell_t'|\le |\hat\ell_t|$.
The last inequality is by Cauchy–Schwartz inequality.

\textbf{Adaptive exploration}:
We first introduce two auxiliary lemmas.
\begin{lemma}
    \label{Lemma::Proof::4::1}
    Let $a_1,\dots, a_T\ge 0$. Then
    \begin{align*}
        \sum_{t=1}^T \frac{a_t}{\sqrt{2\sum_{s=1}^{t-1} a_s+1}}\le 2 \sqrt{\sum_{t=1}^{T} a_t +1}+ \max_{t\in [T]}(a_t).
    \end{align*}
\end{lemma}
\begin{lemma}
\label{Lemma::Proof::4::2}
Given any action sequence $k_1,\dots,k_T$, with the adaptive exploration rate as in Algorithm~\ref{FTRl::general}, there is 
    \begin{align*}
        |\langle \hat\ell'_{t}, \mathbf{c}_t\rangle|\le \ell_{\infty}\Big(2n^2+ \sqrt{2 \sum_{t=1}^{T} |\langle \hat\ell'_{t}, \mathbf{c}_t \rangle|}\Big).
    \end{align*}
\end{lemma}
The detailed proof of Lemma~\ref{Lemma::Proof::4::1} and \ref{Lemma::Proof::4::2} would be provided later.
Now we can prove Lemma~\ref{Lemma::Proof::4}.
    \begin{align*}
        \sum_{t=1}^T \langle \hat\ell'_{t}, \p_t'-\p_t \rangle&\le \sum_{t=1}^T {\rho_t} |\langle \hat\ell'_{t}, \mathbf{c}_t \rangle|\\
        &\le  \sum_{t=1}^T \frac{|\langle \hat\ell'_{t}, \mathbf{c}_t \rangle|} 
   {\sqrt{1+2 \sum_{s=1}^{t-1} |\langle \hat\ell'_{s}, \mathbf{c}_s\rangle|} }\\
        &\le 2\sqrt{1+2 \sum_{t=1}^{T} |\langle \hat\ell'_{t}, \mathbf{c}_t \rangle|}+\max_{t\in [T]}\Big(|\langle \hat\ell'_{t}, \mathbf{c}_t \rangle|\Big)\\
        &\le 2\sqrt{1+2 \sum_{t=1}^{T} |\langle \hat\ell'_{t}, \mathbf{c}_t \rangle|}+\ell_{\infty}\Big(2n^2+ \sqrt{2 \sum_{t=1}^{T} |\langle \hat\ell'_{t}, \mathbf{c}_t \rangle|}\Big).
    \end{align*}
    where the second inequality is due to $\rho_t = 1/(2n^2+ \sqrt{2 \sum_{s=1}^{t-1} |\langle \hat\ell'_{s}, \mathbf{c}_s\rangle|})\le 1/(\sqrt{1+2 \sum_{s=1}^{t-1} |\langle \hat\ell'_{s}, \mathbf{c}_s\rangle|}$, the third inequality is by Lemma~\ref{Lemma::Proof::4::1} with $a_t = |\langle \hat\ell'_{t}, \mathbf{c}_t \rangle|$.
    The last inequality is by Lemma~\ref{Lemma::Proof::4::2}.
    Taking expectation on the both sides, there is 
    \begin{align*}
        \E\Big[\sum_{t=1}^T \langle \hat\ell'_{t}, \p_t'-\p_t \rangle\Big]&\le  \E\Big[2\sqrt{1+2 \sum_{t=1}^{T} |\langle \hat\ell'_{t}, \mathbf{c}_t \rangle|}\Big]+\E\Big[\ell_{\infty}\Big(2n^2+ \sqrt{2 \sum_{t=1}^{T} |\langle \hat\ell'_{t}, \mathbf{c}_t \rangle|}\Big)\Big]\\
        &\le 2n^2\ell_{\infty} + 2\sqrt{1+2 \E\Big[ \sum_{t=1}^{T} |\langle \hat\ell'_{t}, \mathbf{c}_t \rangle| \Big]}+\ell_{\infty}\sqrt{2 \E\Big[ \sum_{t=1}^{T} |\langle \hat\ell'_{t}, \mathbf{c}_t \rangle| \Big]}\\
        &\le 2n^2\ell_{\infty} + 2\sqrt{1+2  \sum_{t=1}^{T} \langle \E\Big[|\hat\ell'_{t}|\Big], |\mathbf{c}_t| \rangle }+\ell_{\infty}\sqrt{2 \sum_{t=1}^{T} \langle \E\Big[|\hat\ell'_{t}|\Big], |\mathbf{c}_t| \rangle}\\
        &\le 2n^2\ell_{\infty} + 2\sqrt{1+2 \sum_{t=1}^{T} \langle |\ell_{t}|, |\mathbf{c}_t| \rangle }+\ell_{\infty}\sqrt{2  \sum_{t=1}^{T} \langle |\ell_{t}|, |\mathbf{c}_t| \rangle  }\\
        &\le 2n^2\ell_{\infty} + 2\sqrt{1+4n \sum_{t=1}^{T} \|\ell_t \|_{\infty} }+2\ell_{\infty}\sqrt{n \sum_{t=1}^{T} \|\ell_t \|_{\infty}  }.
    \end{align*}
    The second inequality is by using Jensen's inequality.
    The fourth inequality is because $\E\Big[|\hat\ell'_{t}|\Big] = |\ell'_t|$ and the magnitude of the truncation loss is not more than that of the original loss, i.e., $|\ell'_t|\le |\ell_t|$.
    The last inequality is due to $\langle|\ell_{t}|, |\mathbf{c}_t|\rangle\le \|\ell_{t}\|_{\infty} \|\mathbf{c}_t\|_1\le 2n \|\ell_{t}\|_{\infty}$. The whole proof is completed.

\subsubsection{Proof of Lemma~\ref{Lemma::Proof::5}}
Recall
\begin{align*}
    \sum_{t=1}^T \langle \hat\ell_{t}-\hat\ell_{t}', \p_t'-\p^\dagger \rangle 
    \le  \sum_{t=1}^T  \|  \hat\ell_{t}-\hat\ell_{t}'\|_1 \le \sum_{t=1}^T  \|\hat\ell_t\|_1 \mathbbm{1}(\hat\ell_{t}\not=\hat\ell_{t}') .
\end{align*}
where the last inequality is due to $\|  \hat\ell_{t}-\hat\ell_{t}'\|_1\le \|  \hat\ell_{t}\|_1$ by the clipping property.
We note that the clipping occurs only if $\hat\ell_t\le 0$ and $\hat\ell_{t,k_t}\le \ell_{t,k_t}/\rho_t$ for every $t\in [T]$ by extra exploration.
Thus,
\begin{align*}
     \sum_{t=1}^T  \|\hat\ell_t\|_1 \mathbbm{1}(\hat\ell_{t}\not=\hat\ell_{t}')\le   \sum_{t=1}^T \frac{|\min(\ell_{t,k_t}, 0)|}{\rho_t}  \mathbbm{1}(\hat\ell_{t}\not=\hat\ell_{t}')\le \frac{\ell_{\infty}^{-}}{\rho_{T+1}}  \sum_{t=1}^T \mathbbm{1}(\hat\ell_{t}\not=\hat\ell_{t}').
\end{align*}
It suffices to prove $\sum_{t=1}^T \mathbbm{1}(\hat\ell_{t}\not=\hat\ell_{t}')\le \log_2(1+\ell_\infty)$.
Notice that $\hat\ell_{t}\not=\hat\ell_{t}'$ will happen if and only if
\begin{align*}
    \ell_{t,k_t}\le 2C_t.
\end{align*}
In this case, we have 
\begin{align*}
        C_{t+1} = 2C_t.
\end{align*}
Now we need to get an upper bound on the size of $C_T$.
In Algorithm~\ref{FTRl::general}, $C_t$ will be updated (i.e., $C_t\not=C_{t+1}$) if and only if the received loss $\ell_{t,k_t}< C_t$.
When $C_t$ is updated, we can note that $C_{t+1}\ge \ell_{t,k_t}$ holds, which also means $|C_{t+1}|\le |\ell_{t,k_t}|$.
Thus, we have
\begin{align*}
    |C_T|\le \max_{t\in [T]}(1,|\ell_{t,k_t}|)\le 1+\ell_{\infty}.
\end{align*}
Since $|C_t|$ is non-decreasing with $t$, it suffices to say that $\ell_{t,k_t}\not=\ell_{t,k_t}^\prime(k_{1:t-1})$ will happen at most $\log_2(1+\ell_\infty)$ times. This completes the proof.

\subsubsection{Proof of Lemma~\ref{Lemma::Proof::3::1}}
 We first note that for every $\alpha\in \R$,
    \begin{align*}
        \arg\min_{\p\in \Delta_n}\Big( \langle L, \p\rangle+\frac{1}{\eta}\Psi(\p)   \Big) =\arg\min_{\p\in \Delta_n}\Big(  \langle L+\alpha\mathbf{1}_n, \p\rangle+\frac{1}{\eta} \Psi(\p)   \Big)
    \end{align*}
    Thus, without loss of generality, we can assume that $L=[L_1,\dots,L_n]^\top$ satisfies
    \begin{align*}
        \sum_{k=1}^n \frac{1}{\eta L_k} = 1;\ L_k\ge 0,\ \forall k\in [n].
    \end{align*}
    Notice that under such conditions, there is 
    \begin{align*}
        \arg\min_{\p\in \Delta_n}\Big( \langle L, \p\rangle+\frac{1}{\eta}\Psi(\p)   \Big) = \arg\min_{\p\in \R^n}\Big( \langle L, \p\rangle+\frac{1}{\eta}\Psi(\p)   \Big)
    \end{align*}
    by KKT conditions.
    
    Now we start the proof.
    By the optimality of $\mathbf{x}$, there is
    \begin{align*}
        \eta L_k + \frac{1}{x_k} = 0,\ \forall k\in [n].
    \end{align*}
    Then we have
    \begin{align*}
        \frac{l}{x_k}\ge -\frac{1}{2\eta x_k} = -\frac{L_k}{2},
    \end{align*}
    thus
    \begin{align*}
        L_k+\frac{l}{x_k}\ge \frac{L_1}{2}.
    \end{align*}
    By the optimality of $\mathbf{x}^\prime$, there exists Lagrangian multiplier $\lambda^\prime$ such that
    \begin{align*}
         \eta L_k -\eta\frac{l}{x_k} +\lambda^\prime  - \frac{1}{x'_k} &= 0,\\
        \eta L_{k'} +\lambda^\prime  - \frac{1}{x'_{k'}} &= 0,\  \forall k'\in [n]\backslash\{k\}.
    \end{align*}
    and satisfies
    \begin{align*}
    \sum_{k'\in [n]\backslash\{k\}} \frac{1}{\eta L_{k'} +\lambda^\prime} + \frac{1}{ \eta L_k +\eta\frac{l}{x_k} +\lambda^\prime}=1.
    \end{align*}
    Using the above, we note that
    \begin{align*}
        x_k' = \frac{1}{ \eta L_k +\eta\frac{l}{x_k} +\lambda^\prime}\le \frac{1}{ \eta\frac{L_k}{2} +\lambda^\prime}.
    \end{align*}
    It suffices to prove that $\lambda^\prime\ge 0$.
    Define function 
    \begin{align*}
        f(\lambda^\prime) = \sum_{k'\in [n]\backslash\{k\}} \frac{1}{\eta L_{k'} +\lambda^\prime} + \frac{1}{ \eta L_k +\eta\frac{l}{x_k} +\lambda^\prime},
    \end{align*}
    we note that
    \begin{align*}
        \sum_{k'\in [n]\backslash\{k\}} \frac{1}{\eta L_{k'} } + \frac{1}{ \eta L_k +\eta\frac{l}{x_k} }\ge \sum_{k=1}^n \frac{1}{\eta L_k }=1,
    \end{align*}
    due to $l\le 0$, which implies $f(0)\ge 1$,
    Since $f$ decreases with $\lambda^\prime$, it suffices to conclude $\lambda^\prime\ge 0$.
    Thus,
    \begin{align*}
         x_k'  \le \frac{1}{ \eta\frac{L_k}{2} +\lambda^\prime}\le \frac{2}{ \eta L_k} = 2x_k.
    \end{align*}
    completes the proof.
\subsubsection{Proof of Lemma~\ref{Lemma::Proof::3::2}}
Similar to the proof of Lemma~\ref{Lemma::Proof::3::1}, it suffices to choose $L=[L_1,\dots,L_n]^\top$ such that
    \begin{align*}
        \eta L_k -\frac{1}{x_k} = 0,\  \forall k\in [n].
    \end{align*}
    By the optimality of $\mathbf{x}^\prime$, there exists Lagrangian multiplier $\lambda^\prime$ such that 
    \begin{align*}
        \eta^\prime L_k +\lambda^\prime -\frac{1}{x'_k} &= 0,\  \forall k\in [n],\\
        \sum_{k=1}^n \frac{1}{ \eta^\prime L_k +\lambda^\prime}&=1.
    \end{align*}
    Similar to the above, it suffices to show that $\lambda^\prime\ge 0$ considering $\eta^\prime\le \eta$.
    Thus, 
    \begin{align*}
        x_k^\prime =  \frac{1}{ \eta^\prime L_k +\lambda^\prime}\le \frac{1}{ \eta^\prime L_k}\le \frac{C}{ \eta L_k} = C x_k.
    \end{align*}
    This completes the proof.

\subsubsection{Proof of Lemma~\ref{Lemma::Proof::4::1}}
    We denote by 
    \begin{align*}
        h_t = \min\Big( \max_{s\in [t-1]}(a_s), a_t \Big),\ b_t = a_t-h_t. 
    \end{align*}
    It suffices to say that 
    \begin{align*}
        \sum_{t=1}^T b_t = \max_{t\in [T]}(a_t).
    \end{align*}
    The proof can be completed as follows.
    \begin{align*}
        \sum_{t=1}^T \frac{a_t}{\sqrt{2\sum_{s=1}^{t-1} a_s+1}}&\le \sum_{t=1}^T \frac{a_t}{\sqrt{\sum_{s=1}^{t-1} a_s+\max_{s\in [t-1]}(a_s) +1}}\\
        &= \sum_{t=1}^T \frac{h_t+b_t}{\sqrt{\sum_{s=1}^{t-1} a_s+\max_{s\in [t-1]}(a_s) +1}}\\
        &\le  \sum_{t=1}^T \frac{h_t}{\sqrt{\sum_{s=1}^{t} h_s +1}}+ \sum_{t=1}^T b_t\\
        &\le 2 \sqrt{\sum_{t=1}^{T} h_t +1}+ \max_{t\in [T]}(a_t)\\
        &\le 2 \sqrt{\sum_{t=1}^{T} a_t +1}+ \max_{t\in [T]}(a_t)
    \end{align*}
\subsubsection{Proof of Lemma~\ref{Lemma::Proof::4::2}}
    \begin{align*}
        |\langle \hat\ell'_{t}, \mathbf{c}_t\rangle| &\le \sum_{k=1}^n \frac{|\ell_{t,k}|\mathbbm{1}({k=k_t})}{p_{t,k}+\rho_t c_{t,k}} |c_{t,k}|\\
        &\le \ell_{\infty}\frac{\mathbbm{1}({k_t^\star=k_t})}{p_{t,k_t^\star}+\rho_t c_{t,k_t^\star}} |c_{t,k_t^\star}|+\ell_{\infty}\sum_{k\in [n]\backslash \{k_t^\star\}} \frac{\mathbbm{1}({k=k_t})}{p_{t,k}+\rho_t c_{t,k}} |c_{t,k}|\\
        &\le \ell_{\infty}\frac{\mathbbm{1}({k_t^\star=k_t})}{1/n-1/2n} n +\ell_{\infty}\sum_{k\in [n]\backslash \{k_t^\star\}} \frac{\mathbbm{1}({k=k_t})}{\rho_t}\\
        &\le \ell_{\infty}\max(2n^2, 1/\rho_t)\\
        &\le \ell_{\infty}\max(2n^2, 1/\rho_{T+1}) = \ell_{\infty}\Big(2n^2+ \sqrt{2 \sum_{t=1}^{T} |\langle \hat\ell'_{t}, \mathbf{c}_t \rangle|}\Big),
    \end{align*}
    where the first inequality is by the definition of $\hat\ell_t'$ and $\p_t'$, the second inequality is by the definition of $\ell_{\infty}$. 
    The third inequality is due to 1). $\p_{t, k_t^\star}$ is one of the largest entries in $\p_t$, which implies $\p_{t, k_t^\star}\ge 1/n$. 2). $c_{t,k_t^\star}\ge -n$ and $\rho_t\le 1/2n^2$ for all $t\in [T]$. 3). $p_{t,k}+\rho_t c_{t,k}\ge \rho_t$ for all $ k\in [n]\backslash \{k_t^\star\}$ by Algorithm~\ref{FTRl::extra_exploration}.
    The last inequality is because $\rho_t$ is nonincreasing.

\end{document}